\documentclass[letterpaper,10 pt, conference]{ieeeconf}  

\IEEEoverridecommandlockouts                              

\usepackage{graphics} 
\usepackage{epsfig} 
\usepackage{mathptmx} 
\usepackage{comment}

\makeatletter
\let\NAT@parse\undefined
\makeatother

\usepackage{amsmath} 
\usepackage{amssymb}  
\usepackage[caption=false]{subfig}
\usepackage{graphicx}
\usepackage{pgfplots}
\usepackage{algorithm}
\usepackage{algpseudocode}
\usepackage{tabularx}
\usepackage{url}

\usepackage{cleveref}
\usepackage{cite}
\usepackage{etoolbox}
\makeatletter
\patchcmd{\@makecaption}
  {\scshape}
  {}
  {}
  {}
\makeatother

\usepackage[letterpaper,bindingoffset=0.2in,%
left=0.75in,right=0.75in,top=0.75in,bottom=0.75in,%
footskip=.25in]{geometry}

\usepackage{accents}

\usepackage{pgfplots}

\newcommand{\executeiffilenewer}[3]{\ifnum\pdfstrcmp{\pdffilemoddate{#1}}{\pdffilemoddate{#2}}>0{\immediate\write18{#3}}\fi}
 
\crefname{equation}{}{}
\Crefname{equation}{Equation}{Equations}
\creflabelformat{equation}{(#2#1#3)}
\crefmultiformat{equation}{(#2#1#3}%
	{--#2#1#3)}{, (#2#1#3)}{ and~(#2#1#3)}

\crefname{figure}{Fig.}{Figs.}
\Crefname{figure}{Figure}{Figures}

\crefname{section}{Sect.}{Sects.}
\Crefname{section}{Section}{Sections}

\crefname{subsection}{Sect.}{Sects.}
\Crefname{subsection}{Section}{Sections}

\crefname{subsubsection}{Sect.}{Sects.}
\Crefname{subsubsection}{Section}{Sections}

\crefname{appsec}{Appendix}{Appendices}
\Crefname{appsec}{Appendix}{Appendices}

\crefname{subappendix}{Appendix}{Appendices}
\Crefname{subappendix}{Appendix}{Appendices}

\crefname{remark}{Remark}{Remarks}
\Crefname{remark}{Remark}{Remarks}

\crefname{prob}{Problem}{Problems}
\Crefname{prob}{Problem}{Problems}

\crefname{constr}{Constraint}{Constraints} 
\Crefname{constr}{Constraint}{Constraints}

\crefname{algorithm}{Algorithm}{Algorithms}
\Crefname{algorithm}{Algorithm}{Algorithms}

\crefname{prop}{Prop.}{Props.}
\Crefname{prop}{Proposition}{Propositions}

\crefname{ALC@unique}{step}{steps}
\Crefname{ALC@unique}{Step}{Steps}

\newcommand{\R}{\boldsymbol{R}}

\title{
\LARGE \bf Interactive Movement Primitives:\\Planning to Push Occluding Pieces for Fruit Picking
\author{Sariah Mghames, Marc Hanheide 
and Amir Ghalamzan E. 
\thanks{The authors are with the University of Lincoln, UK, Centre for Autonomous Systems (L-CAS). This work is partially funded by IUK \#104587, GRASPberry}
}
}

\begin{document}
\maketitle
\thispagestyle{empty}
\pagestyle{empty}

\begin{abstract}
Robotic technology is increasingly considered the major mean for fruit picking. However, picking fruits in a dense cluster imposes a challenging research question in terms of motion/path planning as conventional planning approaches may not find collision-free movements for the robot to reach-and-pick a ripe fruit within a dense cluster. In such cases, the robot needs to safely push unripe fruits to reach a ripe one. Nonetheless, existing approaches to planning pushing movements in cluttered environments either are computationally expensive or only deal with 2-D cases and are not suitable for fruit picking, where it needs to compute 3-D pushing movements in a short time. In this work, we present a path planning algorithm for pushing occluding fruits to reach-and-pick a ripe one. Our proposed approach, called Interactive Probabilistic Movement Primitives (I-ProMP), is not computationally expensive (its computation time is in the order of  100 milliseconds) and is readily used for 3-D problems. We demonstrate the efficiency of our approach with pushing unripe strawberries in a simulated polytunnel. Our experimental results confirm I-ProMP successfully pushes table top grown strawberries and reaches a ripe one.

\end{abstract}

\section{Introduction}
\begin{figure}[tb]\centering
\subfloat[]{\includegraphics[trim=0 3 0 0, clip, width = 0.3\columnwidth]{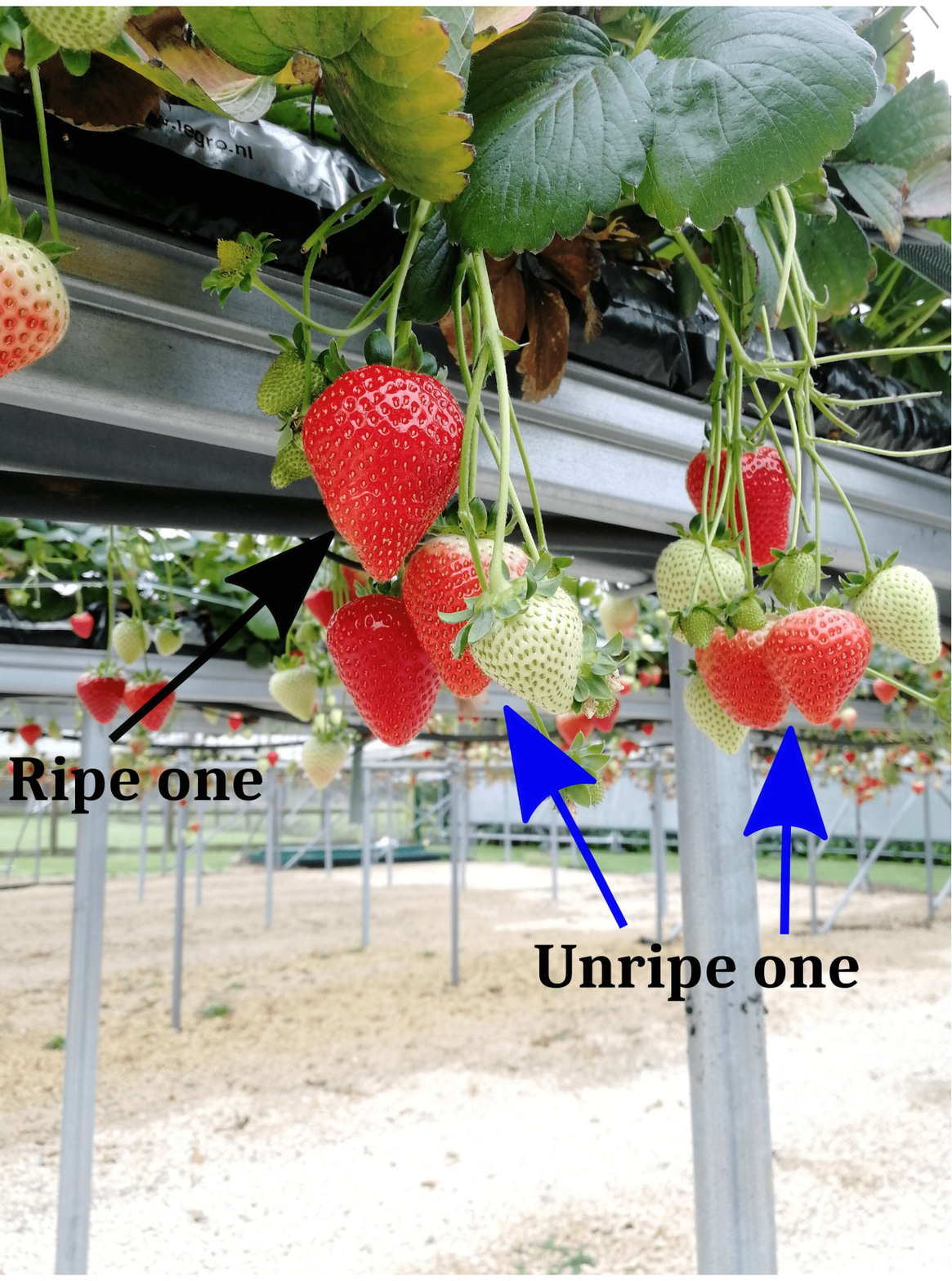}\label{fig:s1}}\hspace{0.01cm}
\subfloat[]{\includegraphics[trim=0 3 0 0, clip, width = 0.3\columnwidth]{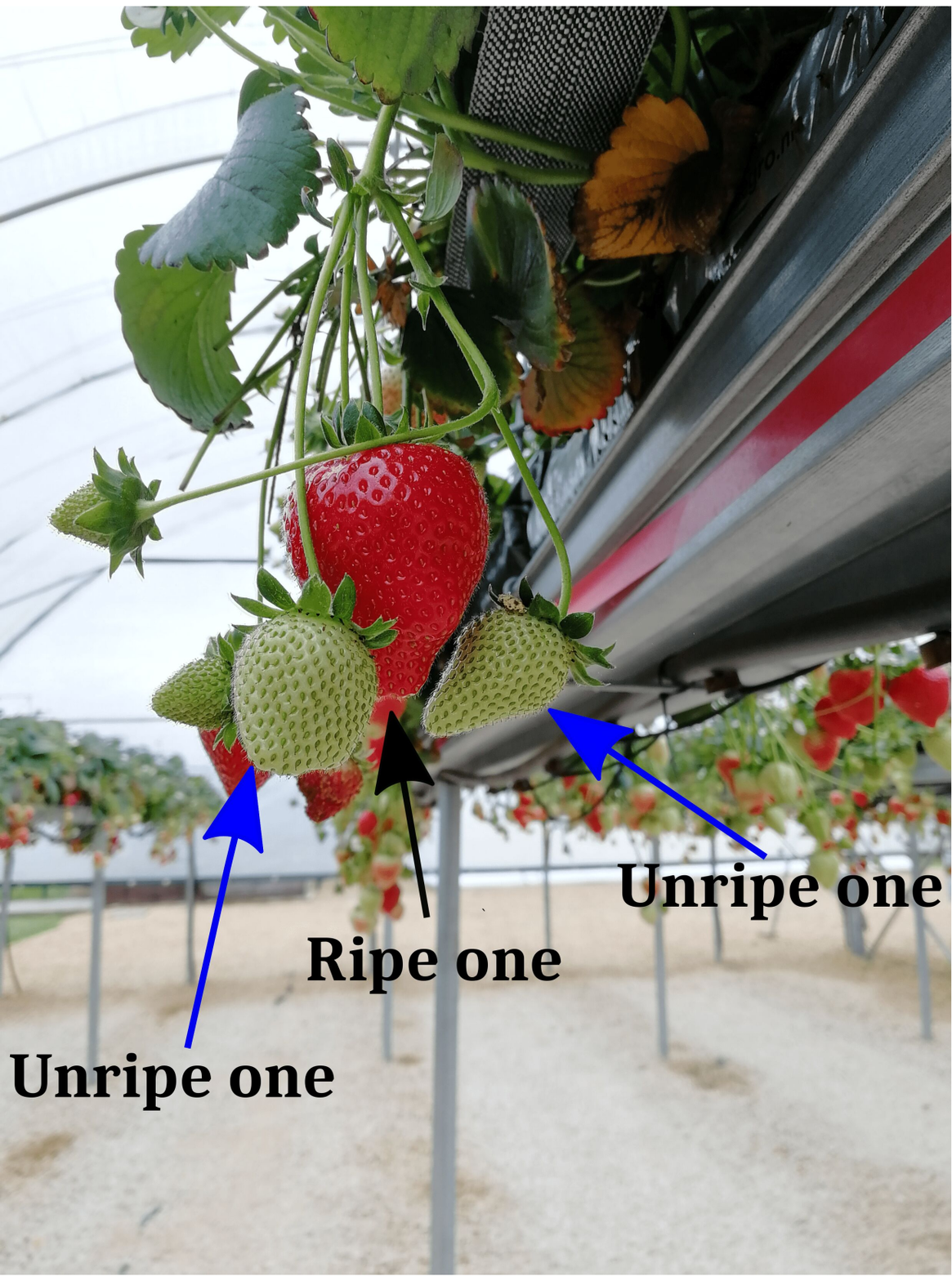}\label{fig:s2}}\hspace{0.01cm}
\subfloat[\label{fig:s3}]{\includegraphics[trim=0 60 0 0, clip, width = 0.35\columnwidth]{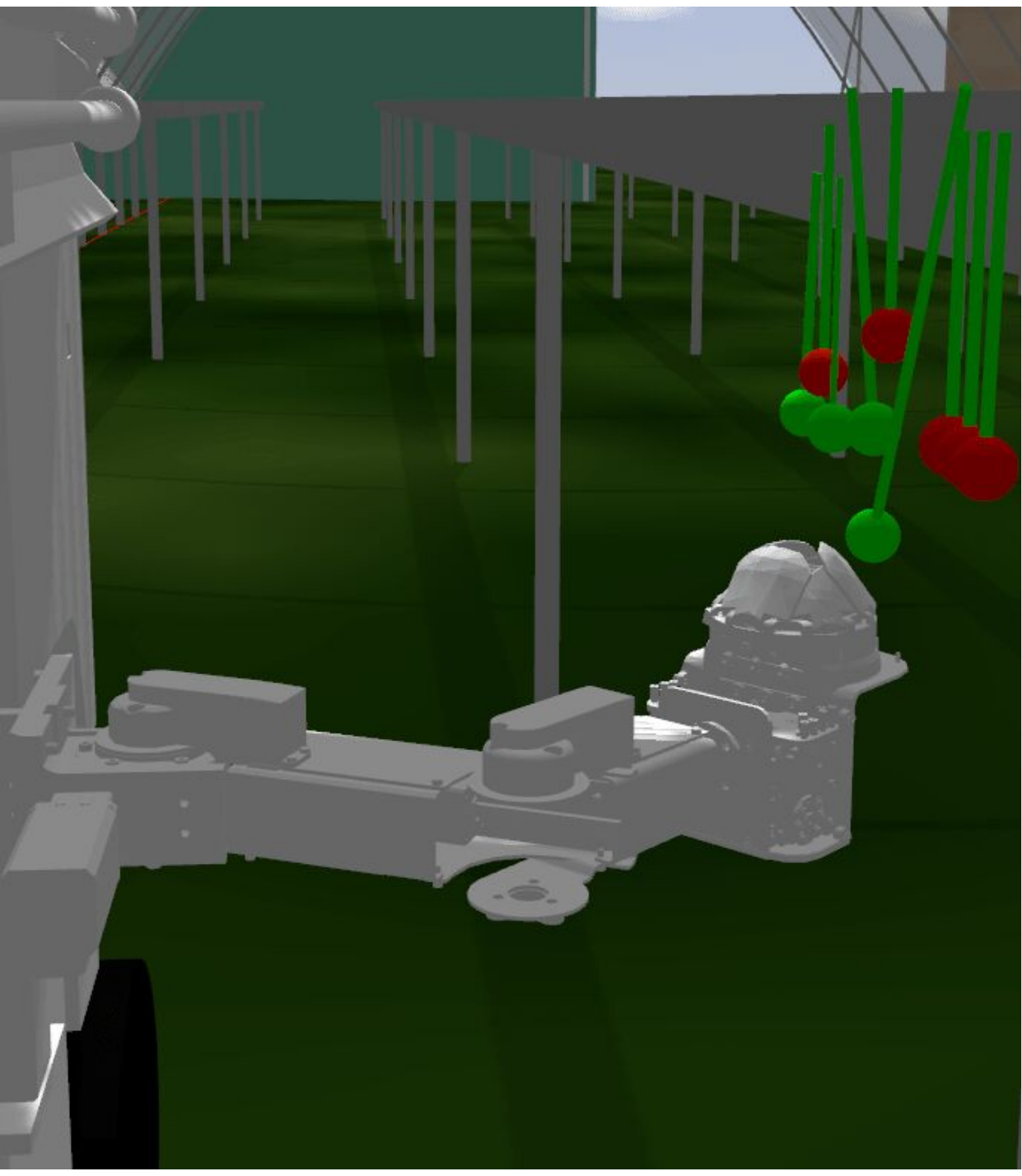}}
\caption{ Table top grown strawberries: \ref{fig:s1} and \ref{fig:s2} show sample clusters of strawberries. The configuration of the strawberries are not uniform or easy to model. Unripe strawberries may occlude the approach path for a strawberry picker robot (\ref{fig:s3}). The table top growing is getting popular because the growing condition can be precisely controlled.~(\ref{fig:s3}) shows our strawberry picking robot consisting of a SCARA arm and a mobile robot. This figure shows the simulation of polytunnel for growing strawberries. We simulate a few possible cluster types important for formalising the pushing movements, similar to the real cluster of strawberries shown in (\ref{fig:s1}). }\label{fig:graphs}
\end{figure}

State-of-the-art 
path planning algorithms do not tackle the problem of fast path generation for a robotic manipulator in a 3-D cluttered scene with connected objects~\cite{ratliff2009chomp, kalakrishnan2011stomp, dogar2012planning, agboh2018real, kitaev2015physics, king2015nonprehensile}. 
%
In this work, we propose an Interactive Movement Primitives (IMP) strategy, that allows us to quickly plan simple quasi-static pushing movements, \emph{e.g.} for fruit picking~\cite{xiong2019autonomous} where the motion planning must readily generalise to different configurations of fruits in clusters (fig.~\ref{fig:graphs}).

\begin{figure}[tb!]
\centering
\subfloat[\label{fig:failure1}]{\includegraphics[trim=0 60 0 0, clip, width = 0.45\columnwidth]{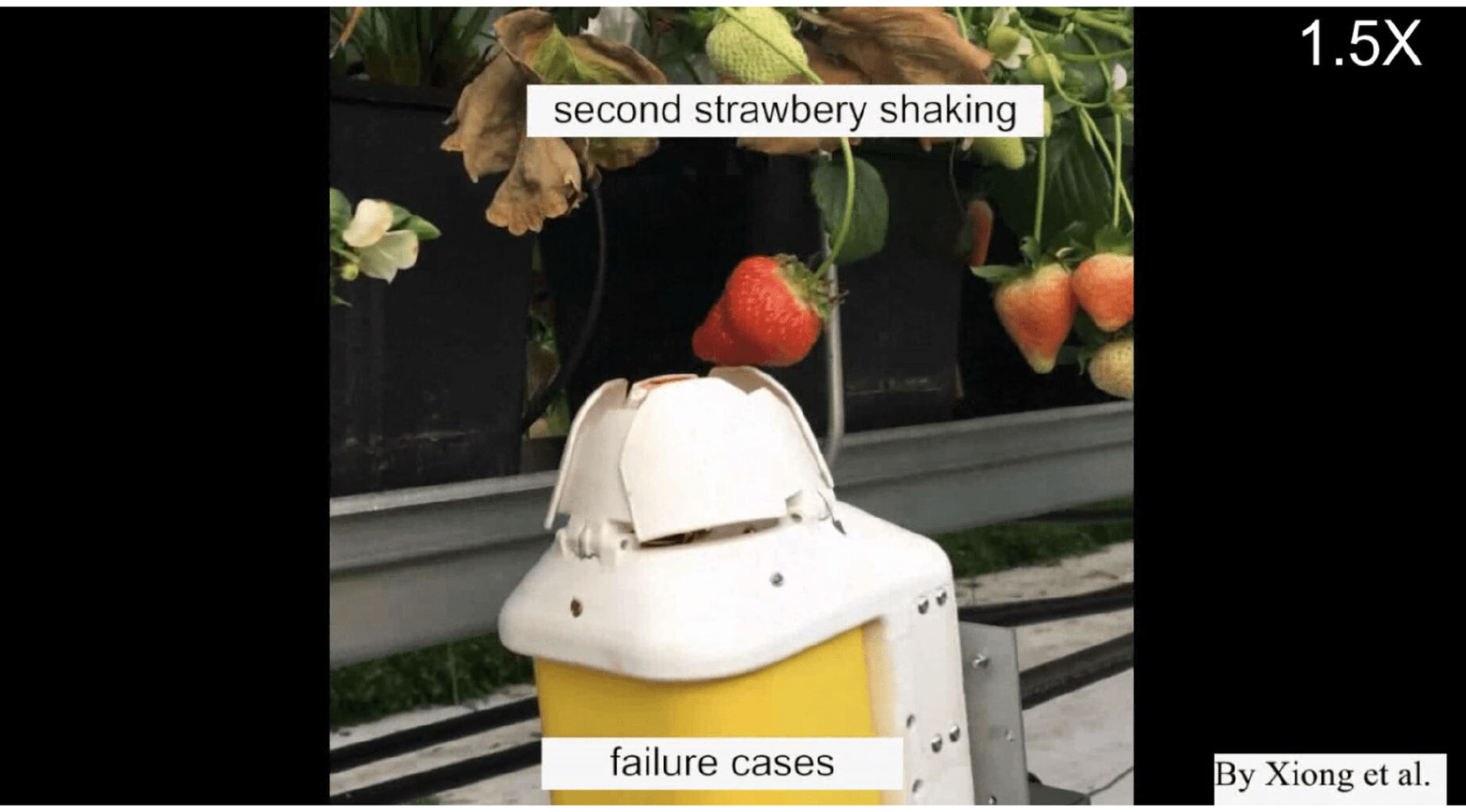}}\hspace{0.2pt}
\subfloat[\label{fig:failure2}]{\includegraphics[trim=0 60 0 0, clip, width = 0.45\columnwidth]{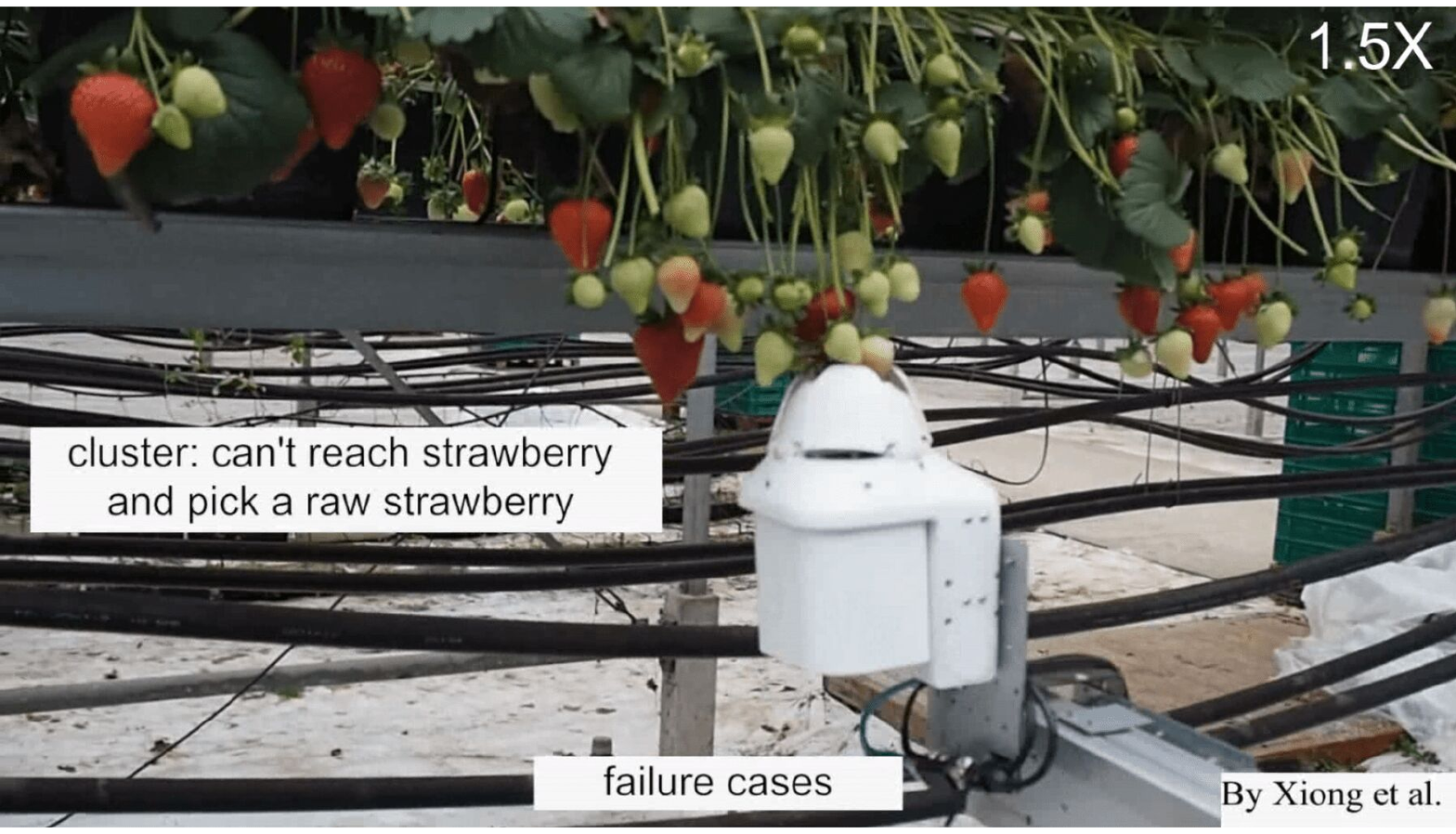}}
\caption{State-of-the-art strawberry picking head~\cite{xiong2019autonomous}: although this sophisticated design can successfully pick most of the isolated strawberries, it fails to pick (i) some of them due to a large error in the planned target position \ref{fig:failure1}, or (ii) all the occluded strawberries~\ref{fig:failure2}. }\label{fig:failures}
\end{figure}

Labour shortage is a major challenge for many sectors, including agriculture. In the UK alone, the soft fruit sector uses 29,000 seasonal pickers to produce over 160,000 tons of fruit every year~\cite{xiong2019autonomous}. 
Only strawberry harvesting costs more than $60\%$ of the total production cost. 
Bringing robotic arms to the field is a response to this global challenge of labour shortage~\cite{Agri}. 
However, precise, reliable and fast motion planning is one of the key bottlenecks of a robotic fruit picker~\cite{xiong2019autonomous}. 
A sophisticated robotic picking technology (fig.~\ref{fig:failures}) is only capable of successfully picking isolated strawberries whereas many of the strawberries are grown in clusters~\cite{xiong2019development}. 

An increasing number of robotic harvesting technologies are nowadays presented. 
Schuetz et al. in \cite{schuetz2015evaluation} formulated the harvesting problem as a static optimal control problem relying on an initially generated heuristic trajectory. 
In their work, the authors generate an optimal harvesting trajectory that minimises the collision and dynamical costs. In \cite{luo2018collision}, the authors presented an energy optimal combined with an artificial potential field approach to formalise the problem of a collision-free path-planning harvesting 6-DoFs robot.
In \cite{xiong2019autonomous}, the authors propose an active obstacle-separation path planning strategy for picking fruits in clusters inspired by human pickers who usually use their hands to push and separate surrounding obstacles during picking. 
In the latter work, the authors adapt a pushing action on the obstacle fruits before reaching the target one. 
The pushing action results mechanically from the move in a designed direction, thanks to a genuine design of a fingers-like gripper. %
Although a pushing action is generated along the path to the target, it doesn't rely on the cluster physics, and hence is considered heuristic and limits the picking success rate. 
In addition, the approach in \cite{xiong2019autonomous} lacks consideration of different kind of occlusions (e.g those coming from stems) which may result in grasp failure due to components coupling, and also, the strategy doesn't consider combined cases (e.g top obstacles on symmetrical side of the target) which can leave the proposed planning strategy without solution.

In addition to picking technology, many other agriculture robotic technologies are researched, e.g. weed detection and removal \cite{tellaeche2008vision}, crops growth monitoring through aerial robots (e.g. using quadcopters)\cite{roldan2016heterogeneous}, fruits \cite{kirk2020b} and plants \cite{yamamoto2016node}, \cite{tabb2017robotic} detection and tracking, mobile robot navigation and mapping \cite{nardi2019actively}-\cite{fentanes20183}. Another agri-robotics area is focusing more on robot kinematics and manipulation trying to find a suitable and efficient gripper design.
Some grippers are developed to achieve stable and soft contacts \cite{elgeneidy2019characterising} with the fruits and others are developed based on a scissor-like concept \cite{xiong2019development}.

%
A human may push/move objects to reach-and-pick a fruit or reach-and-grasp an object in a toolbox or fridge. 
Some previous studies researched some of such problems in robotic context. For instance,~\cite{dogar2012planning, agboh2018real, kitaev2015physics, king2015nonprehensile} consider problems limited to 2-D problem of objects rearrangement on a flat surface, \emph{e.g.} in a fridge or on a shelf, to reach and grasp the desired object in a cluttered scene. 
Although the approach in~\cite{dogar2012planning} is computationally efficient for no-uncertainty case, it is proposed for 2-D occlusion scene. In examples~\cite{agboh2018real, kitaev2015physics, king2015nonprehensile}, computationally expensive approaches, such as physics-based trajectory optimisation, are successfully used. 
However, these approaches require a long computation time for planning the movements (e.g. in~\cite{kitaev2015physics} an average computation time for two objects in a refrigerator is reported for $\sim 10s$).

In other real-world 3-D examples, e.g. picking fruits, the interactions of the objects with its environments may not involve complex computations, \emph{e.g.} friction between object and the flat surface. As such, the approaches listed before are not efficient or readily applicable to our application because they are designed for 2-D problems and need long time for planning and performing pushing movements.

In contrast to optimisation based approaches, robot learning from demonstration (LfD) approaches, \emph{e.g.}~\cite{ragaglia2018robot}, have been successfully developed to minimise the planning time.  
For instance, Dynamic Movement Primitives (DMP) are used to generate and adapt the robot trajectory in real-time~\cite{schaal2006dynamic} where they can also be used to avoid collision~\cite{park2008movement}.
Probabilistic Movement Primitives (ProMP) is also an LfD approach which features interesting properties useful for our problem~\cite{paraschos2013probabilistic} -\cite{paraschos2018using}. 
For example, Maede et al.~\cite{maeda2017probabilistic} proposed an interaction learning method for collaborative and assistive robots based on probabilistic movement primitives. With ProMPs, we are able to encode variability and uncertainty in the movements and to derive new operations which are essential for implementing modulation of a movement, coupling, co-activation and temporal scaling.
Shyam et al.~\cite{shyam2019improving} proposed a probabilistic primitive based optimisation technique to generate smooth and fast trajectories.
Their approach relies on the Covariant Hamiltonian optimisation framework for motion planning with obstacle avoidance constraints initialised with a probabilistic primitive. 

In this paper, we extend ProMP~\cite{shyam2019improving} and propose an Interactive-ProMP (I-ProMP) planning the pushing of unripe strawberries. 
Our contribution is manifold as follows: 
(i) we present primitive cluster types for a challenging strawberry picking problem, defining different configurations of strawberries; 
(ii) we present I-ProMP which efficiently generates, in $0.19s$ mean computation time and $0.0022$ standard deviation, movements necessary for pushing strawberries occluding the robot's way to a ripe one; 
(iii) we develop a simulation environment in \emph{Gazebo 8.0} which allows us to test our I-ProMP. Hence, we test I-ProMP in the developed simulation environment which illustrates our approach successfully performs the pushing movements and reaches the occluded ripe strawberry in different configurations.

The rest of the paper is divided into the following: section \ref{sec:prob} defines the problem and the current main challenge in fruit harvesting, section \ref{sec:approach} presents the approach we followed to tackle the problem, section \ref{sec:exp} illustrates experimental results reported on a simulated field and section \ref{sec:conc} concludes with future works.    
 \label{sec:intro}

\section{Problem Formulation}

\begin{figure}[tb!]
\centering
\setlength{\fboxsep}{1pt}%
\setlength{\fboxrule}{0.5pt}
\begin{center}
\subfloat[\label{fig:cluster1}]{\fbox{\includegraphics[width=4cm,height=2cm]{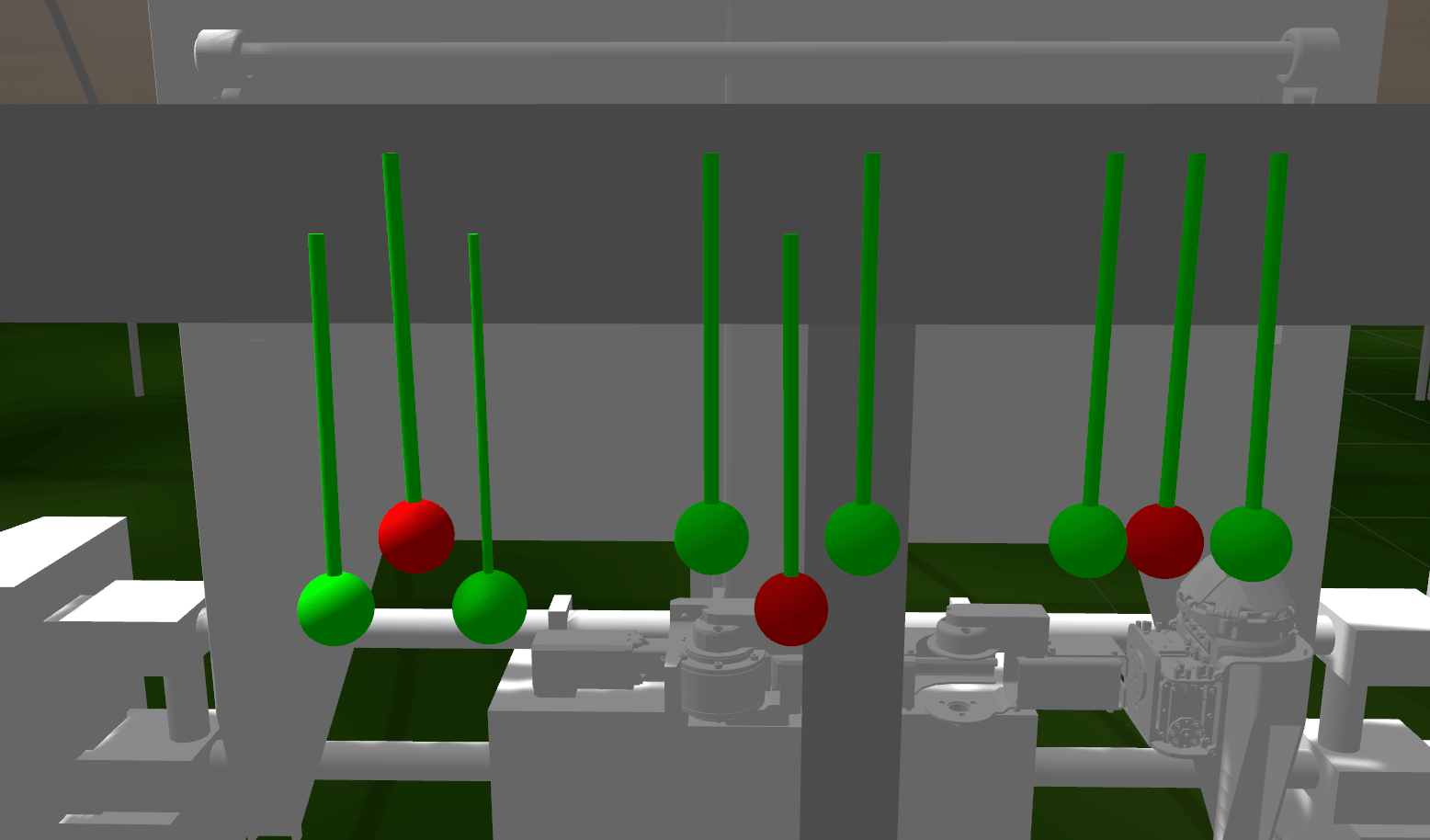}}}\hspace{0.2pt}
\subfloat[\label{fig:cluster2}]{\fbox{\includegraphics[width=4cm,height=2cm]{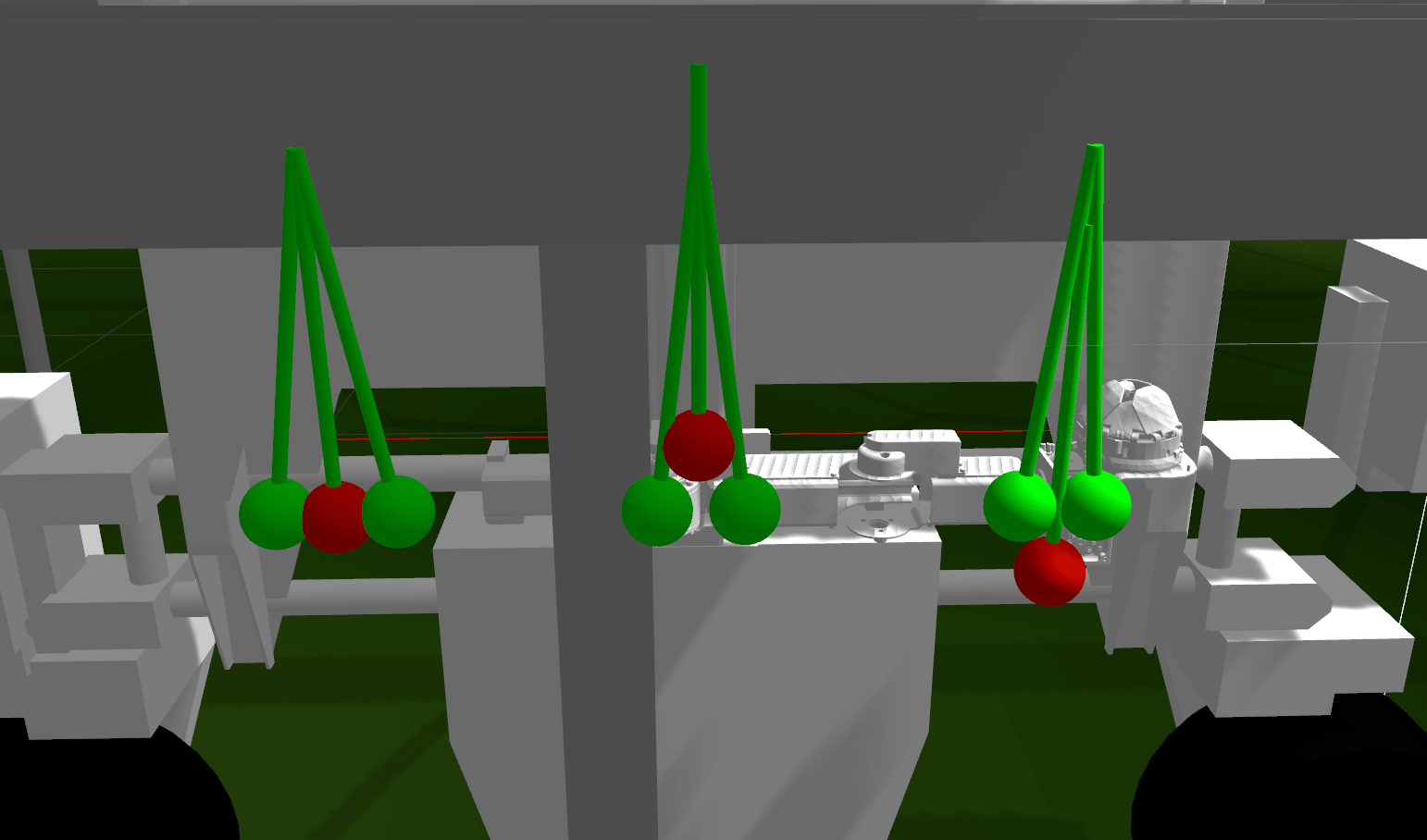}}}
\\
\subfloat[\label{fig:cluster7shot}]{\fbox{\includegraphics[width=4cm,height=2cm]{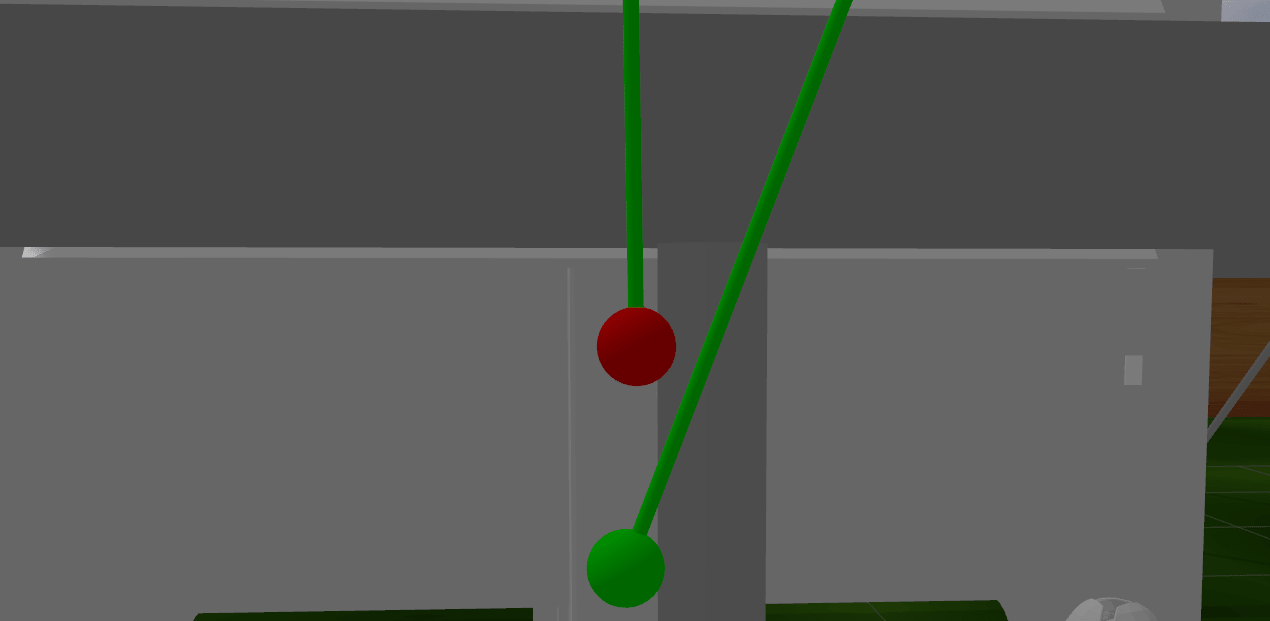}}}\hspace{0.2pt}
\subfloat[\label{fig:cluster10}]{\fbox{\includegraphics[width=4cm,height=2cm]{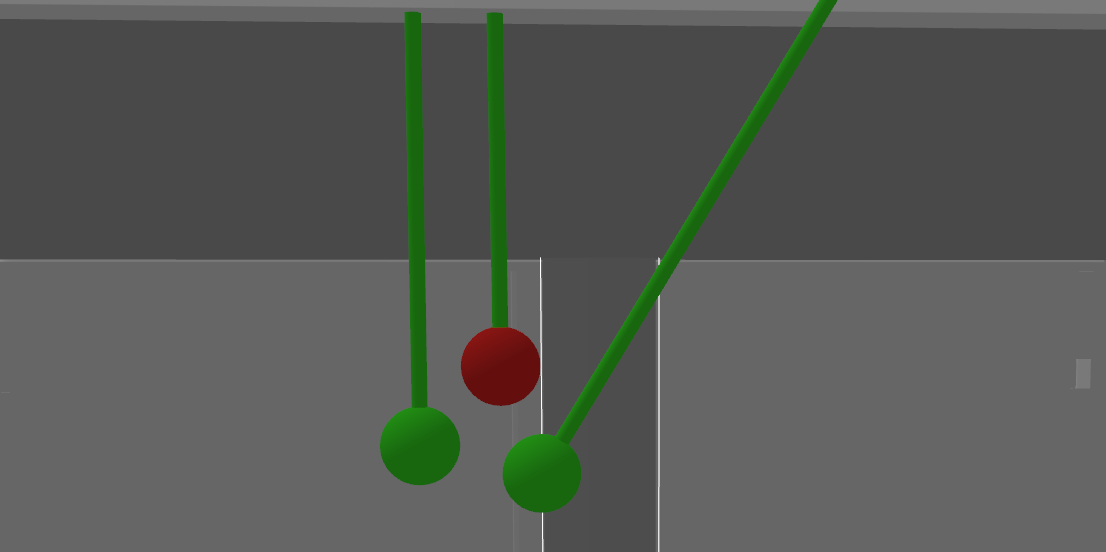}}}
\caption{Cluster configurations identified in this paper and simulated in \emph{Gazebo 8.0}: a target strawberry (red) is surrounded by two neighbours (green) with detached (a) and (b) attached stems lower, $C_I$, upper, $C_{II}$, and at the same level, $C_{III}$ of the target one, from left to right respectively. 
(c) $C_{IV}$: the goal strawberry is occluded by a stiff-inclined stem, unripe strawberry. 
(d) $C_{VI}$: the goal strawberry is occluded by a stiff-inclined stem-strawberry from one side and an unripe (straight) stem-strawberry from the other side.}\label{fig:clusters}
\end{center}
\end{figure} 
Robotic fruit picking is an interesting motion/path planning problem. 
For instance, a ripe fruit to be picked may be located among leaves and unripe fruits where the robot can neither fully observe the fruit nor plan a collision-free robotic arm to reach-and-pick the ripe one. 
Cluster configurations of fruits are determined by the fruit variety which may result in varying number of roots and nodes. 
%
%
The sophisticated end-effector design for strawberry picking~\cite{xiong2019autonomous} fails to pick strawberries in clusters~(fig. \ref{fig:failure2}) because the ripe strawberries may be occluded by unripe ones.
In this paper, we adopt Interactive Probabilistic Movement Primitives to the fruit occlusion problem. 

We identify a few primitive fruit configurations in a cluster illustrated in fig.~\ref{fig:clusters} and fig.~\ref{fig:trajComplexConfig}, which resemble most of the real strawberry clusters. Other cluster types may be formed by combining the primitive configurations\footnote{In all the simulated scene, ripe and unripe strawberries are coloured red and green, respectively}: (i) clusters with isolated and non occluded components (fig.~\ref{fig:cluster1}) -- ripe target above and below and at the same level of unripe neighbours, respectively from left to right; (ii) clusters with connected components -- ripe strawberry at the same level, above and below the neighbour strawberries (fig.~\ref{fig:cluster2}), respectively from left to right; (iii) a cluster with occluded target where the target and occluding strawberry have different nodes (fig.~\ref{fig:cluster7shot}); (iv) a cluster with occluded target-- this configuration shows one of the neighbours occludes the target whereas another does not  (fig.~\ref{fig:cluster10}).  
We simulate all the cluster configurations in \emph{Gazebo 8.0} simulation framework. Each stem is equipped with a 3-axes hinge at its root.  
A discussion for handling each case is elaborated later on in section \ref{sec:exp}. \label{sec:prob}

\begin{figure}
\centering
\includegraphics[width =0.8\columnwidth]{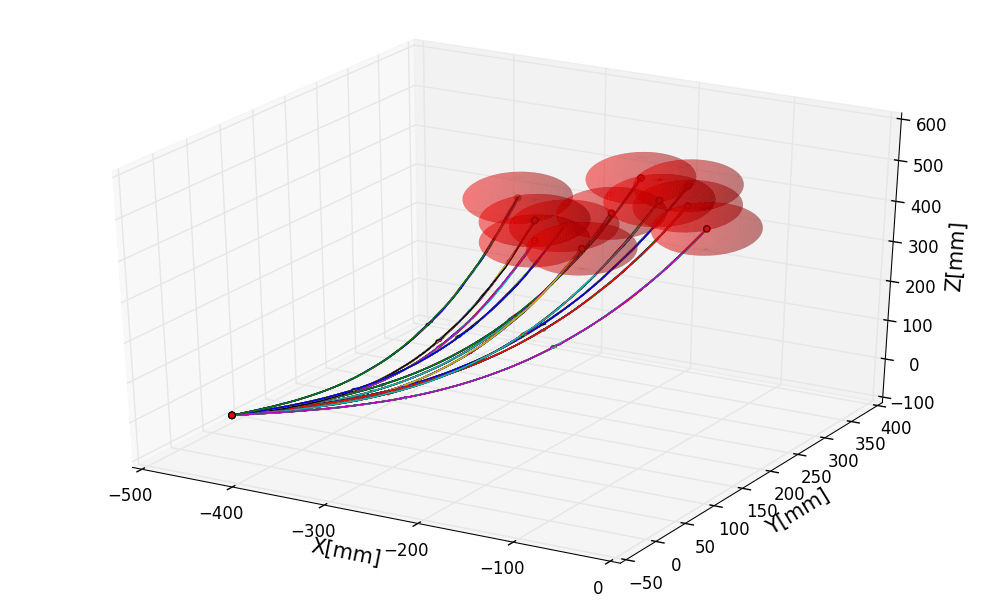}
\caption{Sample paths generated for the picking robotic end-effector. 
We used cubic radial basis functions for generating sample trajectories. Each red sphere resembles a circle surrounding the cluster centred at the goal strawberry. The sample trajectories will be used to train our Probabilistic Movement Primitives (ProMP).}\label{fig:demos}
\end{figure}

\section{Proposed Approach} \label{sec:approach}

In this section, we present our proposed Interactive Probabilistic Movement Primitives (I-ProMP). 
We used cubic radial basis functions for generating sample trajectories, called \emph{demonstrations}, by $\varphi(x) = \|x - c \|^3 $, where x is the input variable and c is a fixed point, called the center of the function. 
We generated 10 sample nominal trajectories (fig.~\ref{fig:demos}) with the same initial and 10 different end-points.  For each nominal trajectory we have 10 samples, their end-points is randomly sampled with the mean equivalent to the nominal goal point and standard deviation  of $10^{-3}$. 
We consider completion time for the demonstrated trajectories to be $T = 1s$.

We model a movement execution as a trajectory $\xi = \{X_t\}_{t=0...T}$, defined by the end-effector pose $X_t$ over time. A ProMP model~\cite{paraschos2013probabilistic} describes multiple ways to execute a movement, which naturally leads to a probability distribution over trajectories. The latter can be represented by a deterministic function of weights $\omega$ and phase variable $z(t)$, as follows: 
\begin{equation}
X_t = \psi_t^T \omega + \epsilon_x 
\end{equation}
where $\psi_t \in \R^{n \times 3}$ is a basis matrix, $\epsilon$ is a zero mean Gaussian random variable with variance $\Sigma_x$. 
We choose k gaussian basis functions which have been shown to be good enough for non-periodic movements, 
\begin{equation} \label{eq:gauss}
\psi_k^G = \exp(-\frac{(z_t - c_k)^2}{2h}) 
\end{equation}
where $z_t$ is a time-dependent phase variable, $c_k$ is the center of the $k^{th}$ basis function and h is the width of the basis. Basis functions in Eq. \ref{eq:gauss} are normalised by $\Sigma_j \psi_j(z)$. 

\subsubsection{ProMP trained by demonstrations}  
In order to learn a movement primitive with properties similar to the generated demonstrations, we learn weight parameters using an extension of the maximum likelihood (\textit{ML}) estimate \cite{bishop2006pattern}, e.g. using the expectation maximization algorithm, as follows
\begin{equation} \label{eq:weight}
\omega = (\lambda I + \psi^T \psi)^{-1} \psi^T X
\end{equation}
where $\lambda$ is a regularisation term to avoid over-fitting in the original optimisation objective~\cite{paraschos2013probabilistic}. 
The probability of observing a trajectory X given the parameter vector $\theta = \{\mu_\omega, \Sigma_\omega\}$ is given by the marginal distribution
\begin{equation}\label{eq:proba}
p(X_t;\theta) = \mathcal{N}(X_t| \psi_t^T \mu_\omega, \psi_t^T \Sigma_\omega \psi_t + \Sigma_x)
\end{equation} 
where $\mu_\omega$ and $\Sigma_\omega$ are the mean and variance of the weight vector respectively. 

\subsubsection{ProMP conditioning at a goal neighbourhood}
We consider a scenario in which the robot has a camera looking at the table from the side view localising a ripe strawberry at $p_{ts}(t) = [x_1,x_2,x_3]$. 
The picking robot first moves to the bottom of the ripe fruit, $p_{bs}(t) = [x_1, x_2, x_3- 0.1]m$, which is captured by the first camera, to get a better view of the cluster with a camera-in-hand. 
Then, the robot performs the push movements to open occlusions and reach the target strawberry. 
We consider the picking actions are cyclic, i.e. after picking strawberry $i$, the robot plans the movements to pick strawberry $i+1$ and so on. 
The picking head of the robot is equipped with a punnet; so, the robot directly picks the strawberries into the punnet. 
As such, target picking position at time $t$ becomes initial position for planning the next picking movements.
At time $t$, we condition the $\mathrm{ProMP}$ at $p_{bs}$ with time $(t+t_1)s$, with $t_1 = 0.85$, and at $p_{ts}$ with time $(t+T)s$.  
In addition, we synthesise a systematic strategy to condition the primitive at selective neighbour fruits, as a first attempt to create a pushing action in an occluded scene.

\begin{figure}
\subfloat[][]{\includegraphics[trim={3.5cm 1cm 1.5cm 2.5cm},clip, width=0.5\columnwidth]{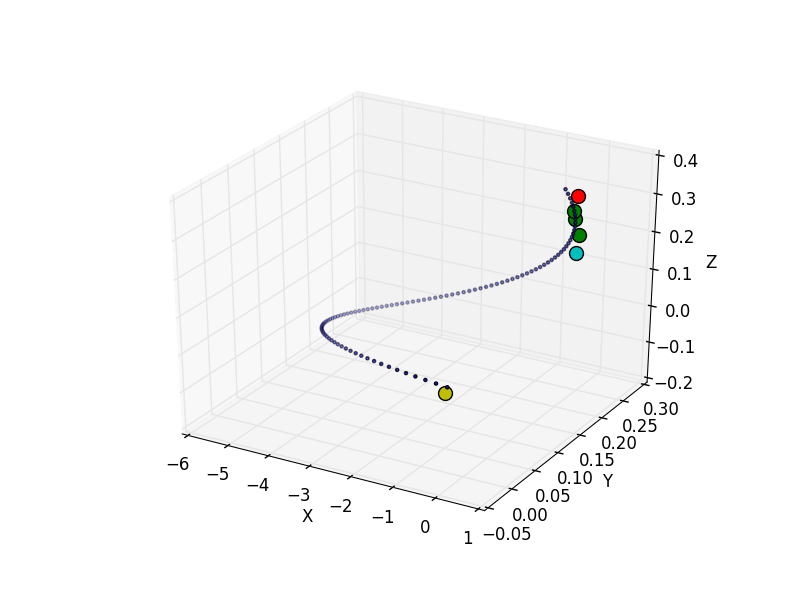}\label{fig:prompB4W1T1}}
\subfloat[][]{\includegraphics[trim={3.5cm 1cm 1.5cm 2.5cm},clip, width=0.5\columnwidth]{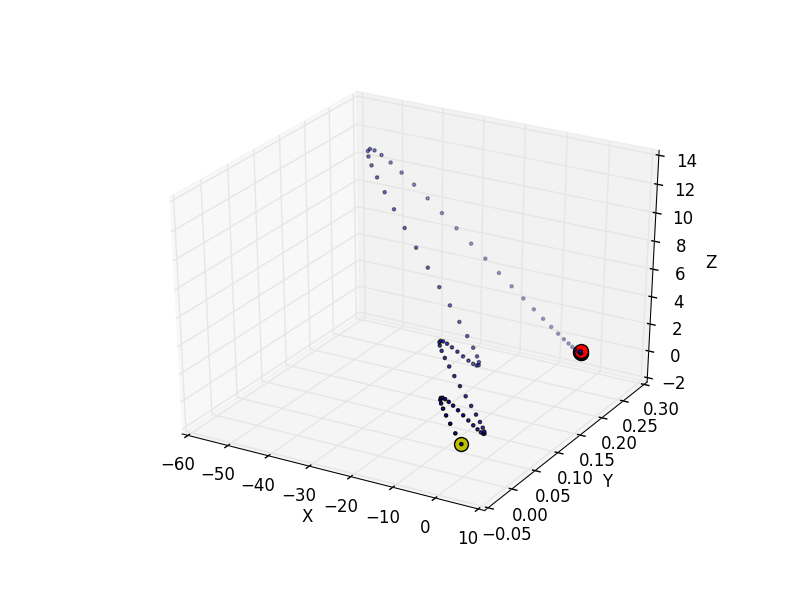}\label{fig:prompB10W1T1}}
\caption{ProMP generation for the same number of simulated fruits: (a) with number of basis $\psi = 4$, bandwidth $h = 1m$ and $T = 1s$; (b) with $\psi = 10$, $h = 1m$ and $T = 1s$. As it these figures show, ProMP with larger number of basis functions can capture more non-linearity of the demonstrated trajectories. }\label{fig:basisT1}
\end{figure}

Fig.~\ref{fig:basisT1} compares the effect of the number of basis functions on the regeneration performance of the learnt ProMP. This figure shows that a ProMP with larger number of basis functions can capture and/or generate higher nonlinear behaviours. 
\begin{figure}
\subfloat[][]{\includegraphics[trim={5cm 1.5cm 2cm 3cm},clip, width=0.16\textwidth]{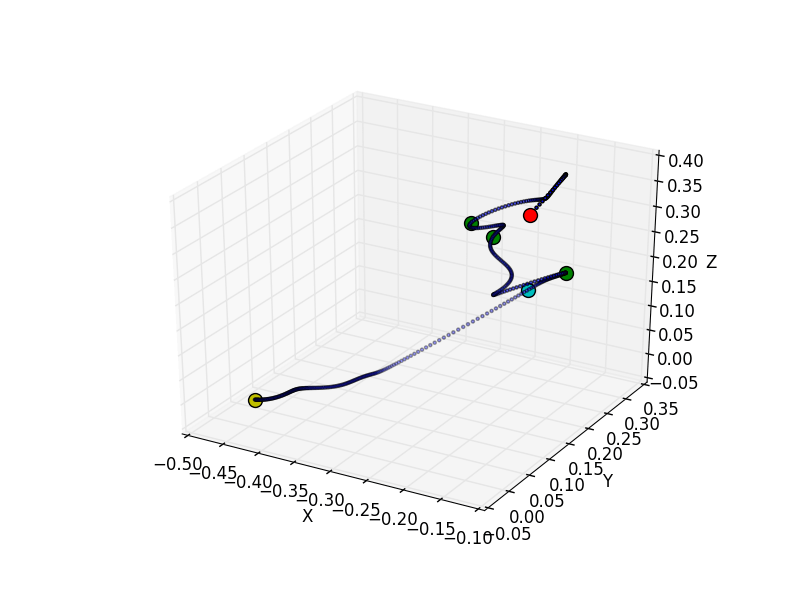}\label{fig:prompC1}}
\subfloat[][]{\includegraphics[trim={5cm 1.5cm 2cm 3cm},clip, width=0.16\textwidth]{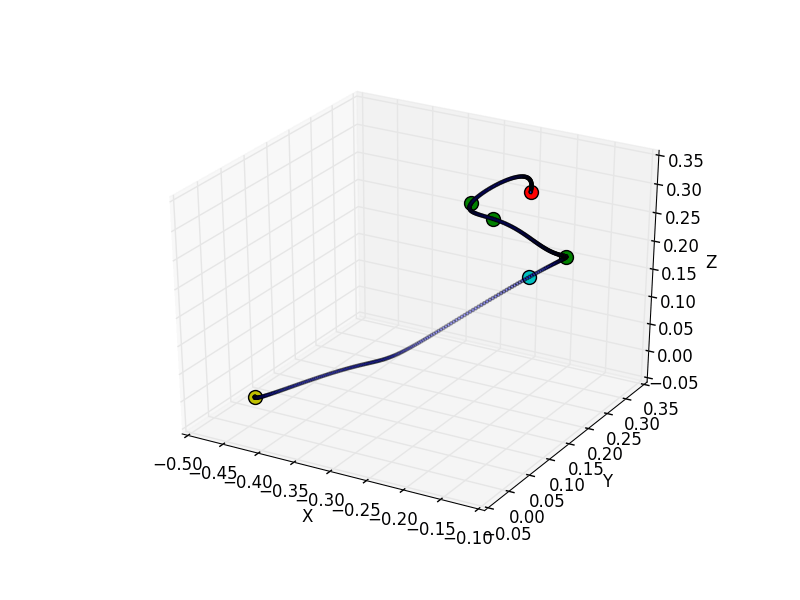}\label{fig:prompC2}}
\subfloat[][]{\includegraphics[trim={5cm 1.5cm 2cm 3cm},clip, width=0.16\textwidth]{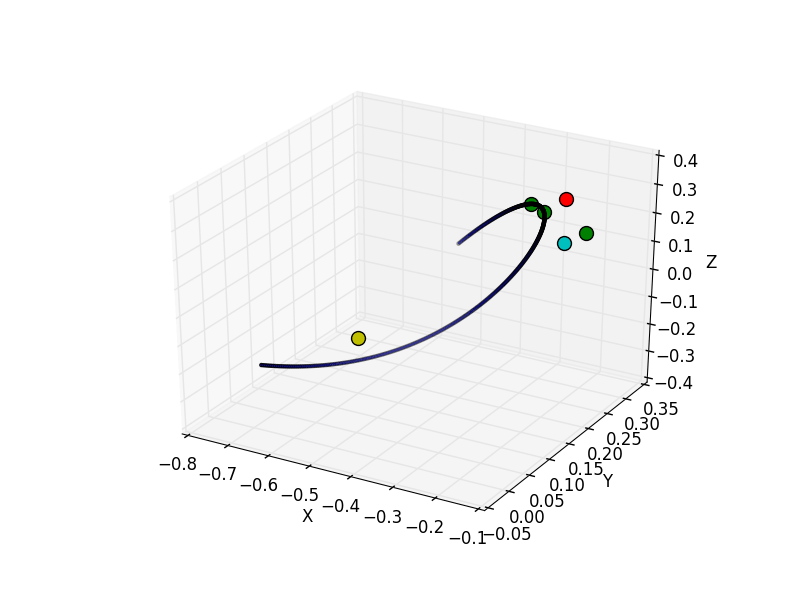}\label{fig:prompC3}} \\
\subfloat[][]{\includegraphics[trim={5cm 1.5cm 2cm 3cm},clip,width=0.16\textwidth]{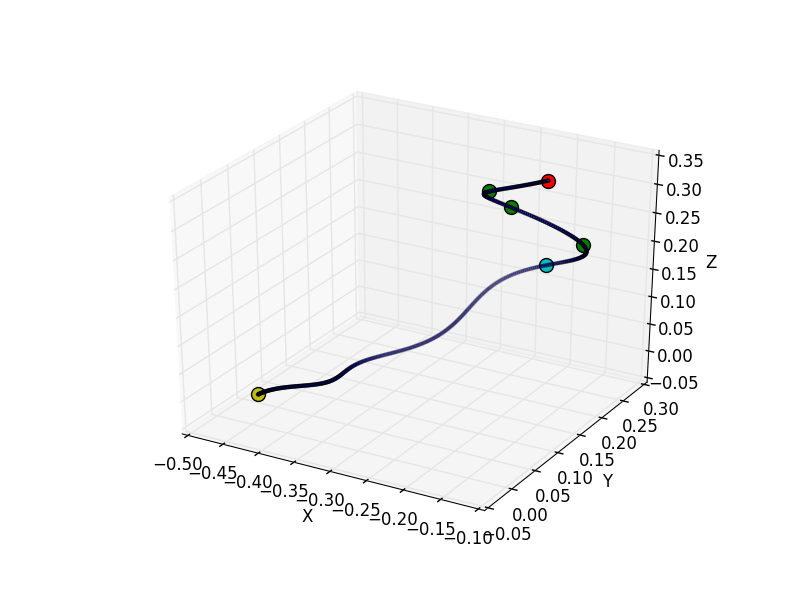}\label{fig:prompC4}}
\subfloat[][]{\includegraphics[trim={5cm 1.5cm 2cm 3cm},clip, width=0.16\textwidth]{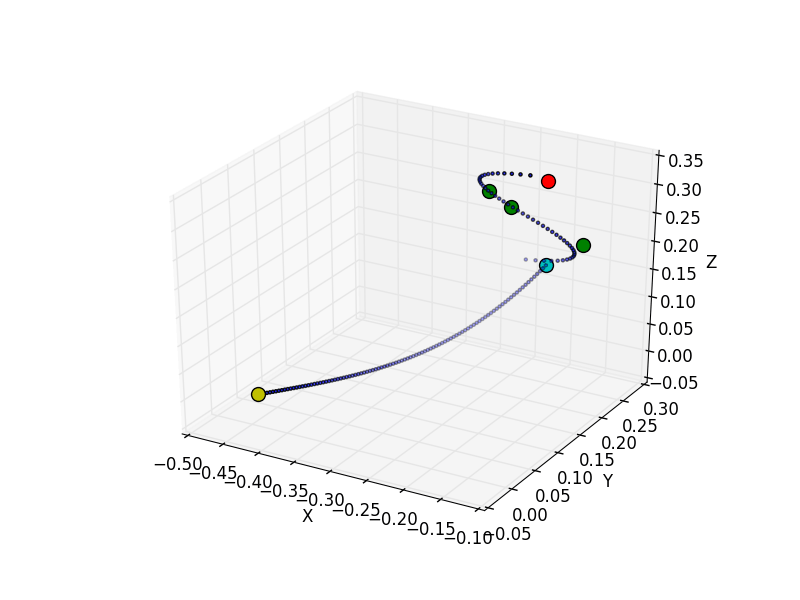}\label{fig:prompC5}}
\subfloat[][]{\includegraphics[trim={5cm 1.5cm 2cm 3cm},clip, width=0.16\textwidth]{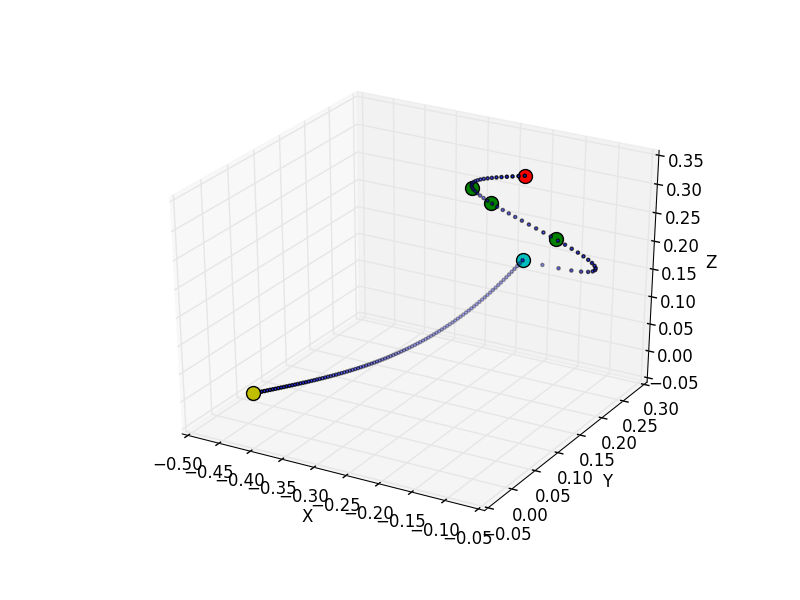}\label{fig:prompC6}}
\caption{(a) One learnt ProMP with number of basis functions $\psi = 20$, bandwidth $h=1m$ (b) $\psi = 10$, $h=1m$ and (c) $\psi = 4$, $h = 1m$, all with conditioning time vector $T_{c1}$. (d) One learnt ProMP with $\psi = 4$ and conditioning time $T_{c2}$, (e) Two learnt ProMPs with $\psi_1= 4$ and $\psi_2=4$, (f) Two learnt ProMPs with $\psi_1 = 4$ and $\psi_2=5$. }\label{fig:basisT2}
\end{figure}


We distinguish an approaching trajectory, which has the least non-linearity, and a pushing trajectory, which may be highly non-linear. 
As such, we consider two ProMPs which represent different level of non-linearities with respect to the phase variable input. The first ProMP, namely $MP_1$ is used to generate reach-to-pick, whereas the second ProMP, namely $MP_2$, is used to generate push-to-pick, as follows:
\begin{equation}
\begin{aligned}
\begin{cases}
\mathrm{MP_1}[\psi_1(z_t)] ,\quad t_0 \leq t < t_1, \\
\mathrm{MP_2}[\psi_2(z_t)] ,\quad t_1 \leq t < T, 
\end{cases}
\end{aligned}
\end{equation}
where, $\mathrm{MP_1}$ and  $\mathrm{MP_2}$ have $k = 4$ and $k = 5$ basis functions, respectively.
Comparing the trajectories generated using $\mathrm{MP_1}$ and  $\mathrm{MP_2}$ models (fig.~\ref{fig:basisT2} e-f) versus using (i) just one ProMP over completion time $T = 1s$  (fig.~\ref{fig:basisT1}) and (ii) just one ProMP over completion time $T = 2s$ (fig.~\ref{fig:basisT2} a-d), shows the superiority of two ProMPs in our specific example. 
%
Fig. \ref{fig:prompB4W1T1} shows that, with a single primitive we are not able to achieve near zero variance ProMP at the conditioned neighbour fruits (unripe) while holding the time duration of the demonstrations, i.e. $T = 1s$. On the other hand, for an increased number of basis functions (e.g $\psi=10$, as in fig. \ref{fig:prompB10W1T1}) a non-linear behavior is induced along the whole trajectory for a 4 desired observations, in addition to the final (ripe fruit) and initial condition (yellow sphere). Hence, we double the total trajectory duration to $T = 2s$ and compare the use of single MP with different number of basis: $\psi=20$ (\ref{fig:prompC1}), $\psi=10$ (\ref{fig:prompC2}) and $\psi=4$ (\ref{fig:prompC3}), all with the same discretised conditioning time $T_{c1} = [0, 0.85, 1, 1.3, 1.6, 2]s$ going from $t_0 = 0s$ to $T = 2s$. The better case turns out to be the one with $\psi=10$. At this point, we compare this case with a similar one, but with a different discretised time duration $T_{c2} = [0, 1.2, 1.4, 1.6, 1.8, 2]s$ (fig. \ref{fig:prompC4}). It turns out that the improvement in smoothness we got in the second phase of the trajectory ($t = 1.2s$ to $T=2s$) results in larger variations in the reach-to-pick trajectory ($t_0$ to $t=1.2s$). As a consequence, we test the learning phase on two different time zones separately while keeping the duration $T=2s$. One ProMP is learnt from demonstrations lasting for $t_1=0.85s$ and the other is learnt for the remaining time duration ($t_1=0.85s$ to $T=1s$). Fig. \ref{fig:prompC5} shows the resulting ProMP with $\psi_{1,2}=4$ while \ref{fig:prompC6} shows the ProMP generated with $\psi_1=4$ and $\psi_2=5$ for the consecutively learnt MPs. 
\vspace{10pt}

Based on the results obtained in fig. \ref{fig:prompC6} that we choose to adopt in this work, it is worth noting that this approach results close to what is addressed by the strategic method in \cite{vijayakumar2000locally} for approximating nonlinear functions in high dimensional spaces. The method in \cite{vijayakumar2000locally} uses a Locally Weighted Projection Regression (\textit{LWPR}) algorithm, where new basis functions are added automatically by the algorithm when needed.


Given the desired end-effector observation $\xi^*_t = [ X^*_t, \Sigma_x^*]$ with a desired variance $\Sigma_x^*$, we can get the Gaussian conditional distribution $p(\omega^*|\xi^*_t)$ for the updated weight vector $\omega^*$, with mean and variance the maximum a-posteriori estimate (\textit{MAP}), as follows 
\begin{equation}\label{eq:mean}
\mu_\omega^* = \mu_\omega + \Sigma_\omega \psi_t (\Sigma_x^* + \psi_t^T \Sigma_\omega \psi_t)^{-1} (X_t^* - \psi_t^T \mu_\omega), 
\end{equation} 
\begin{equation}\label{eq:var}
\Sigma_\omega^* = \Sigma_\omega - \Sigma_\omega \psi_t (\Sigma_x^* + \psi_t^T \Sigma_\omega \psi_t)^{-1} \psi_t^T \Sigma_\omega
\end{equation} 

In a second phase, an online decision-making on the neighboured fruits to condition upon is made by a planner presented in the following, based on the criteria $(\mathbf{i} \cdot \mathbf{d}_{ng}) \leq r_g^{max}$, where $\mathbf{d}_{ng}$ is the euclidean distance between a neighbour fruit and the target one, projected on unit vector along the table top axis $\mathbf{i} = [1, 0, 0]$. $r_g^{max} = D_{d, max}/2$ is the maximum gripper opening radius, $3cm$ .

\subsection{\textit{Interactive}-ProMP}
In the following section, we present a \textbf{S}tochastically-driven \textbf{I}nteractive \textbf{P}lanner (SIP) which implements a probabilistic physics-based pushing strategy to swallow a target fruit from an 3-D cluster.

\begin{figure}
\centering
\includegraphics[trim={5cm 3cm 3cm 4cm},clip,width=0.8\columnwidth]{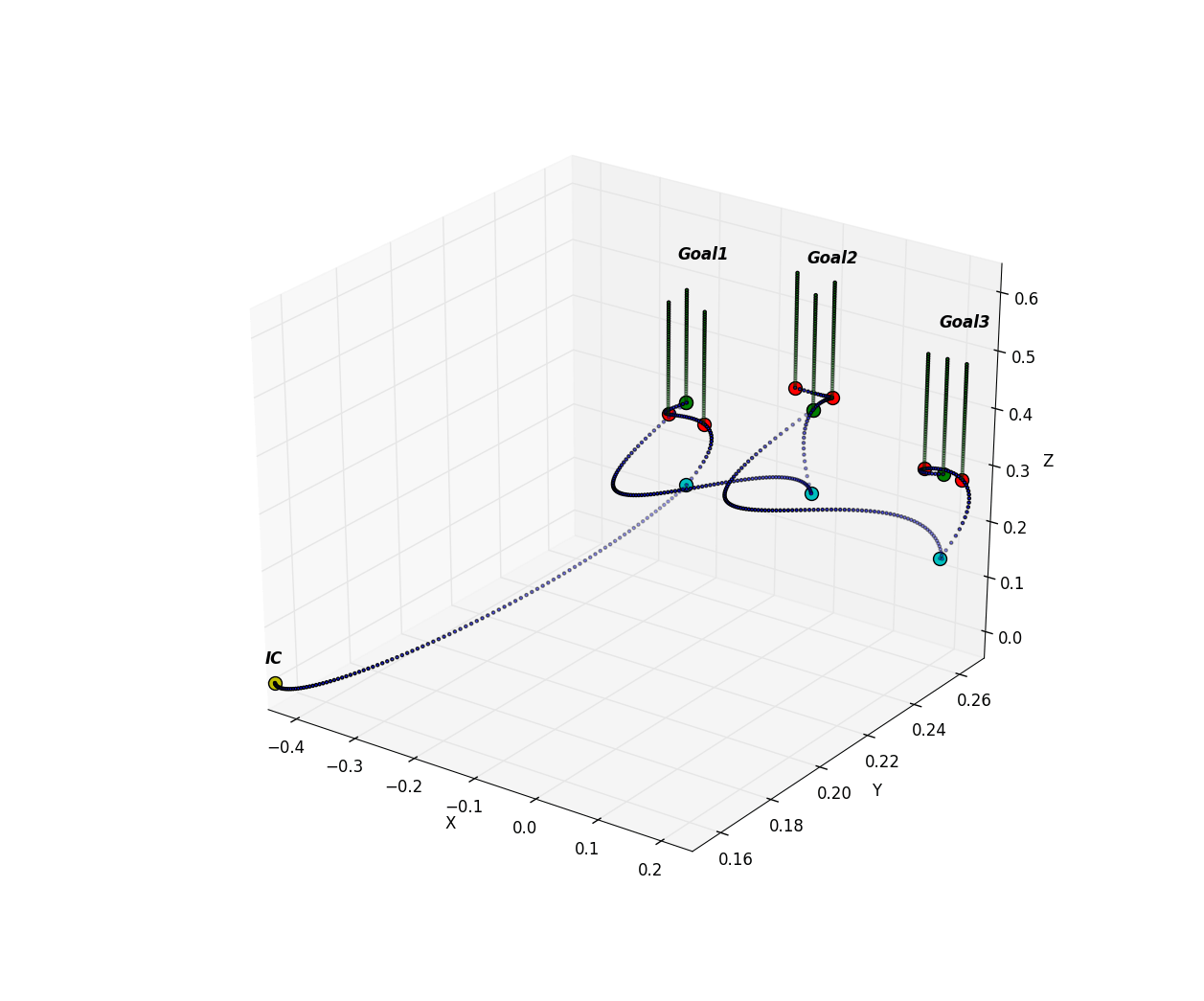}
\caption{Sequential probabilistic primitives generated from an initial state (IC) to \textit{Goal1} fruit (green sphere) while conditioning on its neighboured fruits (red spheres) as a first attempt to generate a pushing action, then from \textit{Goal1} to \textit{Goal2} followed by \textit{Goal2} to \textit{Goal3}. The 3 primitives are illustrated for 3 types of clusters shown in fig. \ref{fig:cluster1}.
Two task space learnt probabilistic movement primitives are conditioned at a distance $10cm$ (cyan spheres) below the goal for a close fruit detection.}\label{fig:cond_promp}
\end{figure}

With the \textit{ProMP} generated above (Eqs. \ref{eq:proba}, \ref{eq:mean}, \ref{eq:var}) and illustrated in fig. \ref{fig:cond_promp}, two problems can still arise depending on the cluster configuration in scene:
\begin{itemize}
\item The \textit{ProMP} may cross the stem or the target fruit before being able to swallow it. Consequently, a damage to the scene and a large error in the goal pose can arise.
\item The neighboured fruit by which the \textit{ProMP} passes may return to its natural position by physics effect.
\end{itemize}
In order to reduce the aforementioned risks and increase the swallow success rate, we present hereafter a modified version of the conditioned movement primitive that features the following:
\begin{itemize}
\item Decision making on the number of neighboured fruits that can free the robot path to the goal. 
\item Decision making on the local pushing direction of a  neighboured fruit
\item Decision making on the amount of pushing, represented by an estimate of the angle of rotation of the stem
\end{itemize}

\begin{figure}
\centering
\subfloat[][]{\includegraphics[width = 3cm, height = 3.7cm]{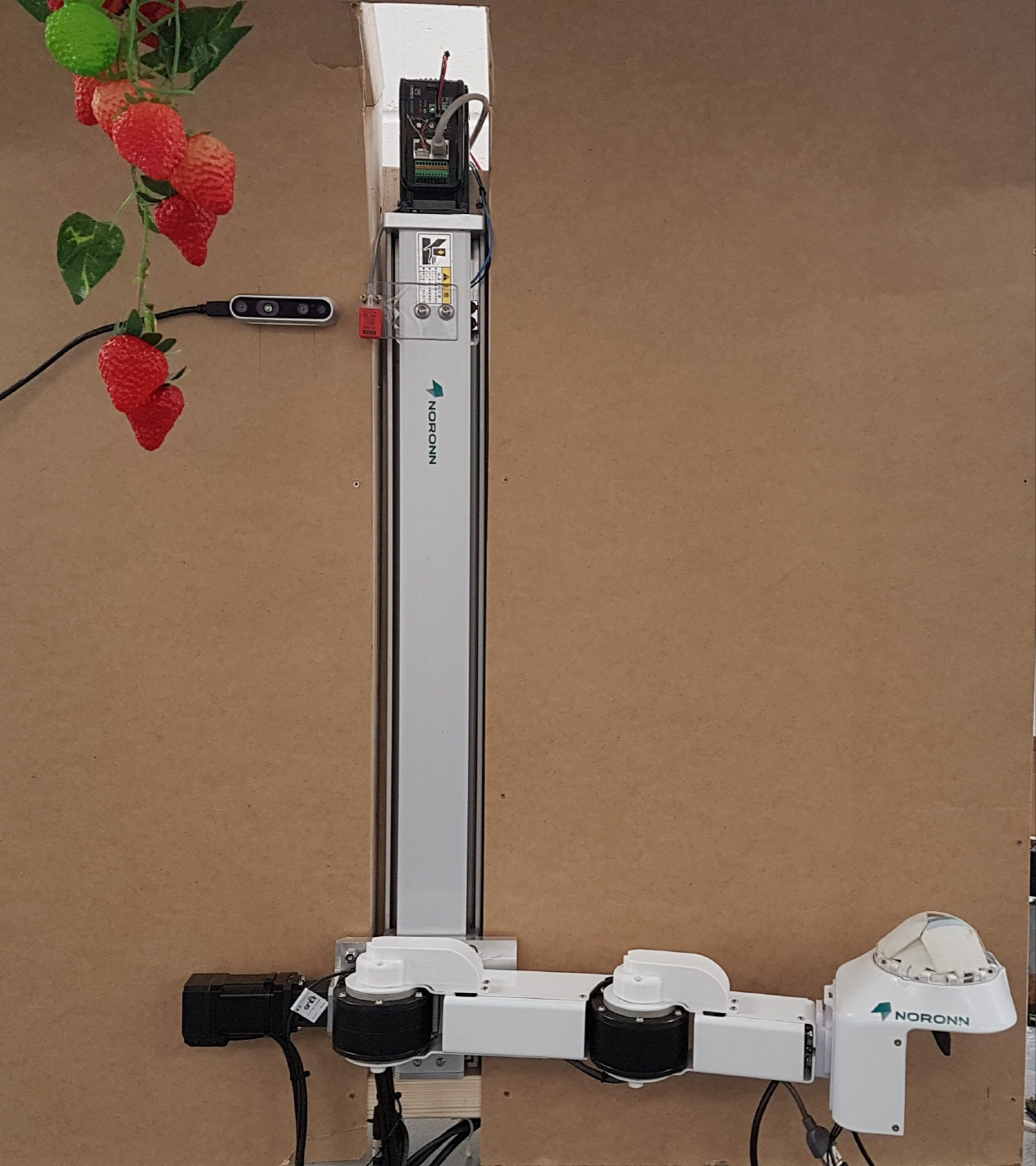}\label{fig:mock}} \hspace{25pt}
\subfloat[][]{\includegraphics[trim=2 40 2 2, clip, width=0.38\columnwidth]{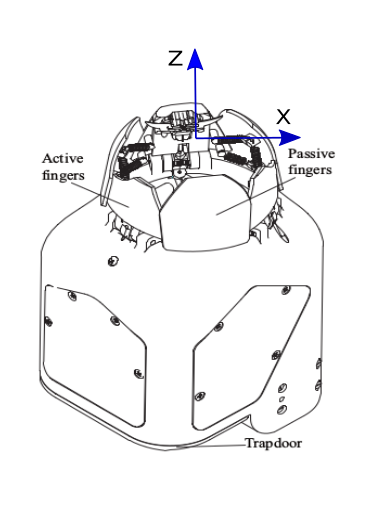}\label{fig:gripper}}
\caption{(a) Mock-up setup with SCARA arm, a finger-like gripper, an RGB-D camera and a fake strawberry cluster, (b) Finger-like gripper \cite{xiong2019autonomous} integrated with a scissor and 3 infra-red sensors to localise the target inside. }\label{fig:hardware}
\end{figure}

\begin{algorithm}[t]
\caption{The 3-D Pushing Strategy}\label{euclid1}
\hspace*{\algorithmicindent} \textbf{Input:}
$\mathbf{G}$ = \{matured fruits\}, $\mathbf{\Bar{G}}$ = \{non-matured fruits\}\\
\hspace*{\algorithmicindent} \textbf{Output:} $P^{c_t}_p$, $P^{c_t}_o$, $P_p^{c_{t}^*}$ 
\begin{algorithmic}[1]
\Procedure{Pushing nearest neighbours}{}
\State $g_t \gets random(\mathbf{G}) $
\State $\mathbf{u}_{tt} \ni (\mathbf{u}_{tt} \in table\,top)  \cap  (\mathbf{u}_{tt} \parallel table\, top) $
\State $C^{n_t}_p \gets RNN( g_t, \mathbf{G}_s = \mathbf{G} U \mathbf{\Bar{G}}, r = 0.05)$
\State $S_d, S_p, S_t \gets Subset(C^{n_t}_p, \mathbf{d_{n_t g_t}}, \mathbf{u_p})$

\State $ n \in \{order(S_d, axis=\mathbf{k})\}$
\State $S_d^* \gets SubsetOPT(S_d, S_p, S_t)$
\For {$n_d^* \in S_d^*$} 
\State $root = \mathbf{u}_{stem} \cap \mathbf{u}_{tt}$
\State $ L_{stem} = \mathbf{d}_{root,n_i}$
\State $\theta_0 = \arccos(\mathbf{u}_{stem} \cdot \mathbf{k})$
\State $\theta = \arcsin(r_g^{max} / L_{stem})$
\State $d\theta = \theta - \theta_0$ 
\State $\mathbf{s} \gets L_{stem}^2 + L_{stem}^2 - 2*L_{stem}^2*\cos(d\theta)$
\State $\mathbf{u}_p \gets GetDir(\mathbf{v}_1, \mathbf{v}_2, S^*_d)$
\State $P_p^{c_{t}^*} \gets Proj(\mathbf{u}_p, \mathbf{s})$
\EndFor
\EndProcedure
\end{algorithmic}
\end{algorithm}

\vspace{5pt}
\textit{\textbf{The SIP algorithm:}} the SIP is presented in two stages. Algorithm \ref{euclid1} takes as input a set of ripe $\mathbf{G}$ and unripe $\mathbf{\Bar{G}}$ fruits and output the position of the pushable objects selected at time t ($P_p^{c_t}$), their orientation at time t ($P_o^{c_t}$) estimated by the stem orientation (assumed given in this work), their updated position at t ($P_p^{c_{t}^*}$). At first, the algorithm selects a target $g_t$, generates a cluster $C_p^{n_t}$ associated to $g_t$ using the radius nearest neighbours technique (\textit{RNN}) where $r=0.05m$ is the chosen cluster radius. The clustered points are then divided into bottom, plane and top subsets, $S_d$, $S_p$ and $S_t$ respectively, minimised by the criteria ($\mathbf{i} \cdot \mathbf{d}_{n_t g_t}) \leq r_g^{max}$. 
$S_d$ is ordered incrementally along axis $\mathbf{k}=[0, 0, 1]$ (operation 6), then optimised for the number of elements (operation 7). The optimization takes into account the points in the subset that are at quasi-equal level and selects the one ($S_d^*$) with smaller ` $\mathbf{i} \cdot \mathbf{d}_{n_t g_t}$' value, in case their stems doesn't occlude the target, else (e.g fig. \ref{fig:cluster10}) it chooses the one that complies more with the gravity direction because it needs less effort to push it. The latter fact is not taken as a showcase in this work, related results will be reported in our future work when an optimization problem formulation will be elaborated for that purpose. Given the stem orientation $\mathbf{u}_{stem}$, the algorithm computes a stem length estimation, $L_{stem}$, for every element in the optimised subset, i.e $n_d^*$, after getting the intersection of $\mathbf{u}_{stem}$ with the table top plane (operation 9), then gets the inclination angle $\theta_0$, computes the total inclination $\theta$ needed to free the maximum gripper opening (operation 12), and gets the minimum displacement required ($\mathbf{s}$) of the pushable object to free the path to the target. Two possible pushing directions are proposed, $\mathbf{v}_1$ and $\mathbf{v}_2$, each normal to the stem direction $\mathbf{u}_{stem}$ (assumed to be vertical, hence aligned with $\mathbf{k}$, only for the case of determining the pushing direction $\mathbf{u}_p$). If the algorithm finds other elements from $S^*_d$ at same level as $n_d^*$ (e.g fig. \ref{fig:cluster10}), it pushes $n_d^*$ along $\mathbf{v}_2$ (i.e normal to table top), else, it pushes $n_d^*$ along $\mathbf{v}_1$ (i.e parallel to table top). The updated $n_d^*$ coordinates (i.e $P_p^{c_{t}^*}$) are computed by projecting the displacement vector $\mathbf{s}$ on the pushing direction $\mathbf{u}_p$ (operation 16).

\begin{algorithm}[t]
\caption{Interactive ProMP generation}\label{euclid2}
\hspace*{\algorithmicindent} \textbf{Input:} $g_{t-1}, g_t, P^{c_t}_p$, $P_p^{c_{t}^*}$\\
\hspace*{\algorithmicindent} \textbf{Output:} Interactive ProMP 
\begin{algorithmic}[2]
\Procedure{ProMP}{}
\State $\mathbf{P}_{cond} = \{g_{t-1}, P^{c_t}_p, P_p^{c_{t}^*}, g_t \} $
\State $\omega_{ML} \gets Eq. (3)$
\State $\mu_\omega , \Sigma_\omega$
\For {$ wpt \in \mathbf{G}_{cond}$} 
\State $x_d , \Sigma_d$
\State Update: $\mu^*_\omega, \Sigma^*_\omega \gets Eq. (6), (7)$
\EndFor
\EndProcedure
\end{algorithmic}
\end{algorithm}

Algorithm \ref{euclid2} takes as input the latest ripe goal $g_{t-1}$, a new ripe goal $g_t$, $P_p^{c_t}$ and $P_p^{c_{t}^*}$. It then conditions the learnt ProMPs at the input set (operation 5) and updates the parameters of the weight distribution (operation 7).
\vspace{5pt}

\textit{Discussion:} Clusters of type \textit{Goal2} and \textit{Goal3} in fig. \ref{fig:cond_promp} are not considered at this stage of work by SIP, hence $S_p$ and $S_t$ are eliminated in algorithm \ref{euclid1}.  In a future work, we will benefit from the genuine finger-like design of the gripper in use (shown in fig. \ref{fig:gripper}) to test the grasp of corresponding scenarios with pushing actions generated kinematically by an incremental opening and closing of the active fingers. 

\section{Experiments and Results} \label{sec:exp}
\begin{figure}
\centering
\setlength{\fboxsep}{1pt}%
\setlength{\fboxrule}{0.5pt}
\subfloat[\label{fig:traj2}]{\includegraphics[trim={4cm 1cm 1cm 3cm},clip, width=0.5\columnwidth]{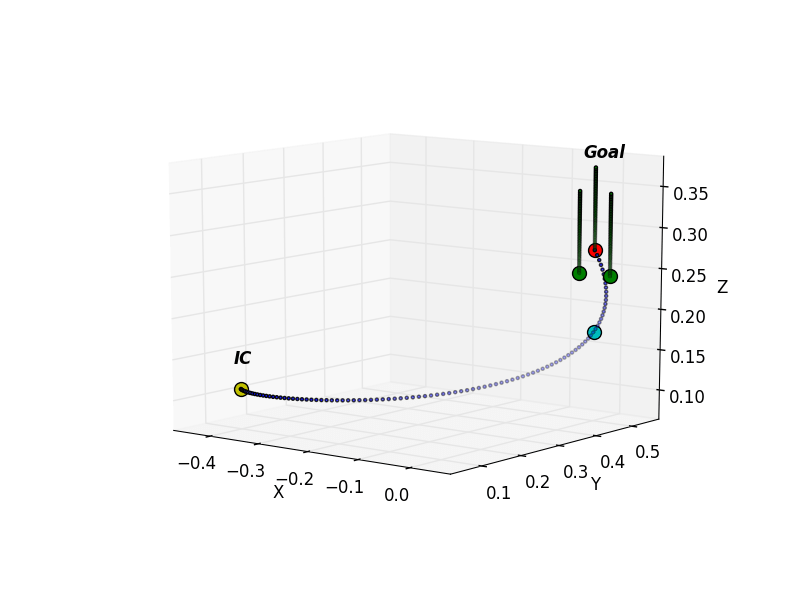}}
\subfloat[\label{fig:traj_sim2}]{\includegraphics[trim={3.5cm 1cm 1cm 2cm},clip, width=0.5\columnwidth]{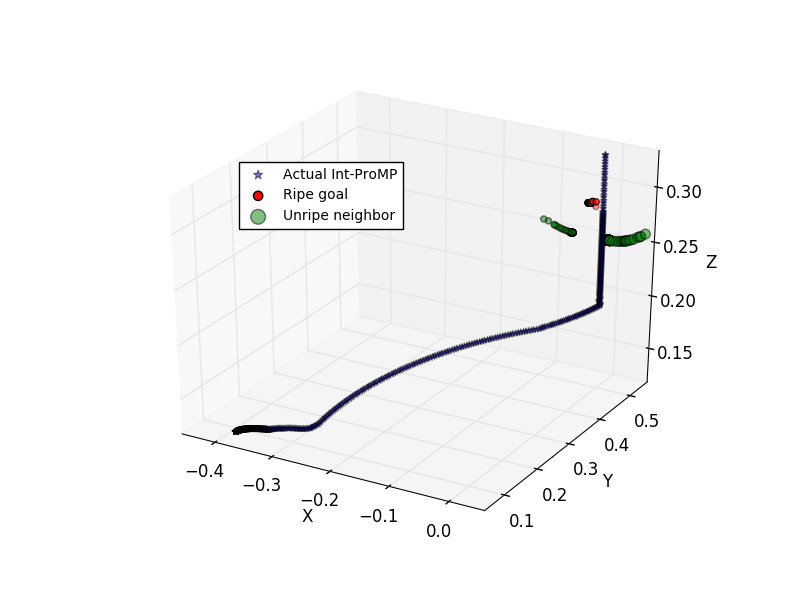}}
\caption{Probabilistic primitive-based strategy to swallow a soft fruit from unoccluded clusters: no pushing actions are needed}\label{fig:trajNormConfig}
\end{figure}

Given a finger-like gripper design \cite{xiong2019development} that we adopt in our application, we consider that configurations of type \ref{fig:cluster1} and \ref{fig:cluster2} can be targeted with only a primitive-based planner applied on the ripe target. A pushing action will result mechanically from path following.
This hypothesis is tested for the case in which the target is above neighbours and results are reported in fig. \ref{fig:trajNormConfig}, where fig. \ref{fig:traj2} represents a ProMP generated by the planner and fig. \ref{fig:traj_sim2} represents the simulated end-effector trajectory (in blue), unripe fruits trajectory and target ripe fruit trajectory. 

For the case of fig. \ref{fig:cluster10}, as discussed in section \ref{sec:approach}, this type of cluster gets optimised for the number of neighbours to push and gets a pushing direction along $\mathbf{v}_2$, i.e normal to table top.
In the following, we present results related to the configuration in fig. \ref{fig:cluster7shot} and other two descendants.

\subsection{Hardware Platform}
Two SCARA arms are mounted on Thorvald mobile robot \cite{grimstad2018software} for fruit harvesting. Each arm is a 3-DoFs PRR\footnote{P: prismatic, R: revolute} serial chain and consists of a cable-driven fingers-like end-effector whose role is to swallow a strawberry, center it, and finally cut it with an internal scissor. For a detailed description of the gripper the reader can refer to \cite{xiong2019development}. 
Whenever the goal fruit is detected by 3 integrated infra-red sensors, the end-effector $X_z$ position is increased by $2cm$ to cut the stem with the scissor. As the growing season starts in June \cite{Agri}, we only present our results in a simulated polytunnel.

\begin{figure}[!h]
\subfloat[\label{fig:sim1}]{\includegraphics[trim={0cm 4.5cm 0cm 0cm},clip,width =0.49\columnwidth]{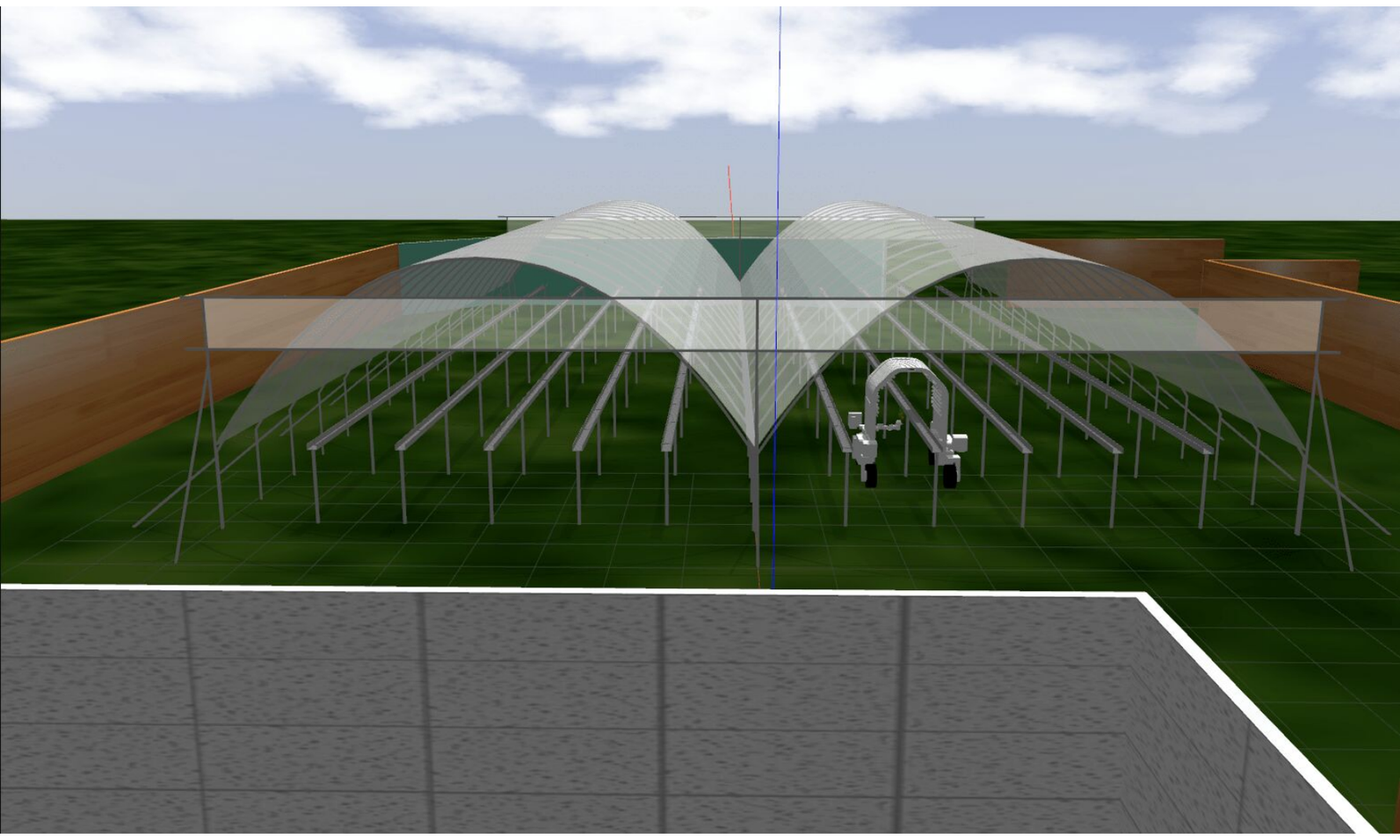}} \hspace{2pt}
\subfloat[\label{fig:sim2}]{\includegraphics[trim={4.5cm 4cm 0.06cm 4cm},clip,width =0.473\columnwidth]{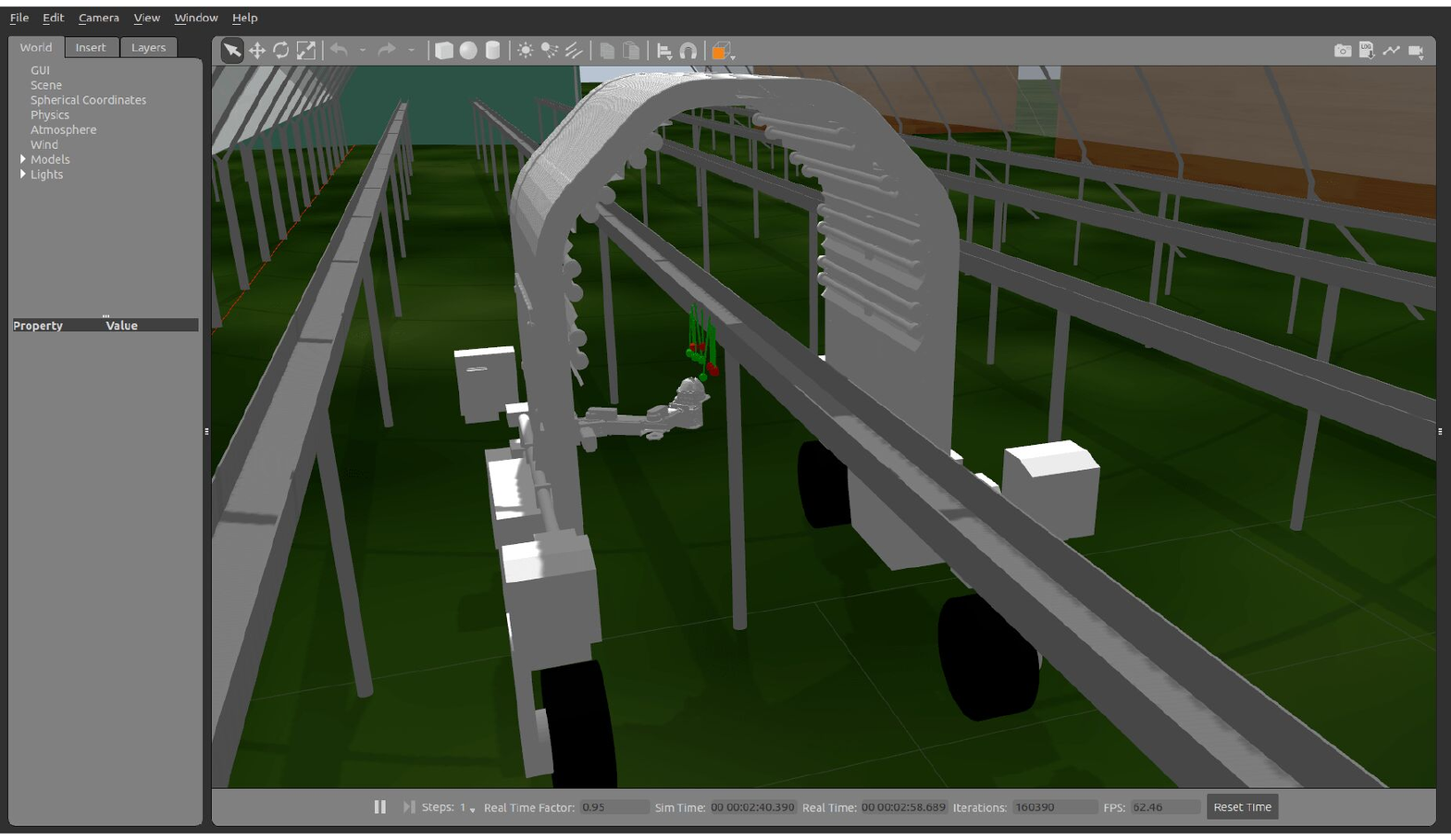}}
\caption{Field simulation on the SCARA arm mounted on a mobile base, the Thorvald robot (b), navigating in polytunnels (a)}
\end{figure}

\subsection{Simulations and Results}
\begin{figure*}[t]
\centering
\setlength{\fboxsep}{1pt}%
\setlength{\fboxrule}{0.5pt}
\subfloat[\label{fig:cluster7}]{\fbox{\includegraphics[trim={20cm 10cm 20cm 10cm},clip, width=0.2\textwidth]{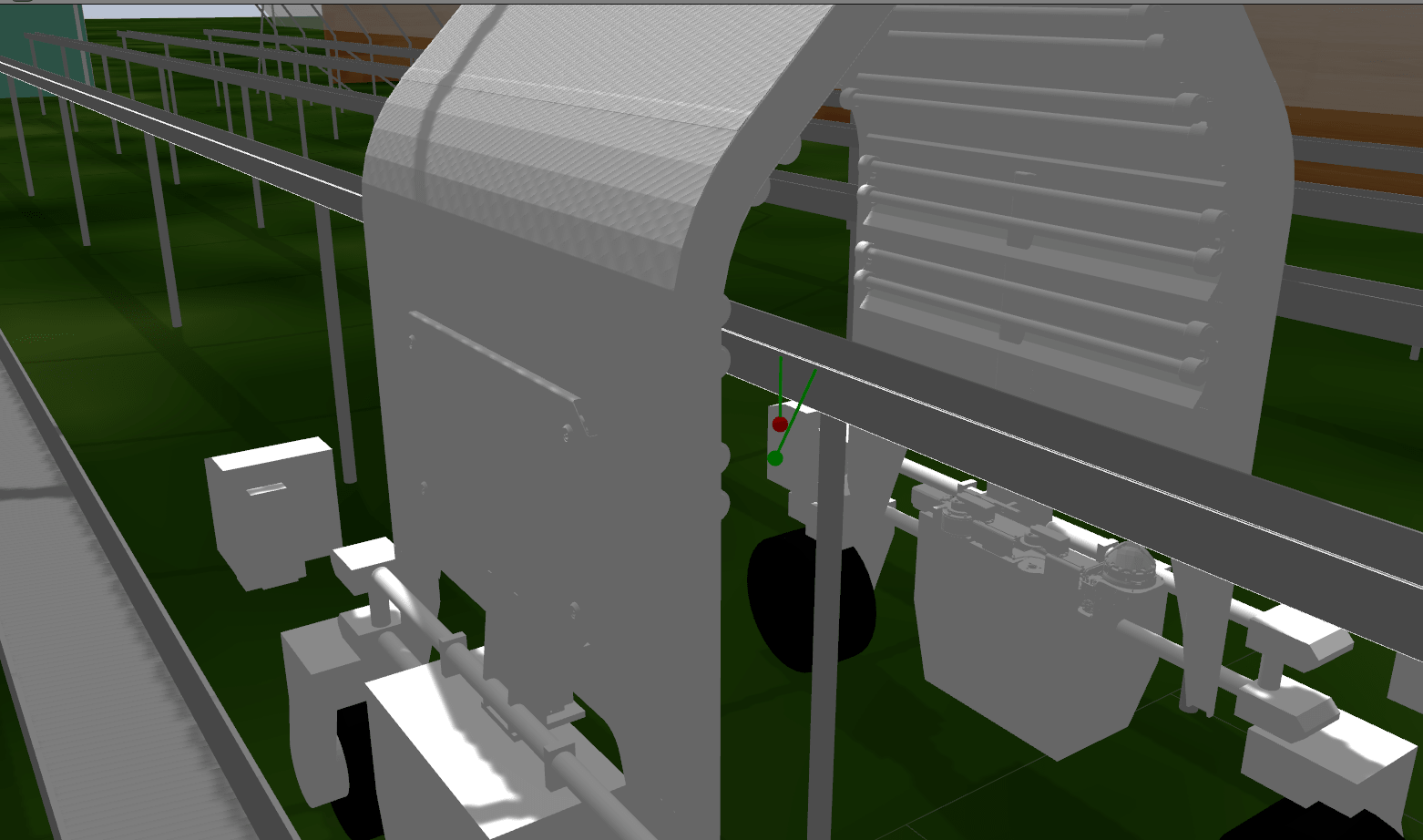}}}\hspace{50pt}
\subfloat[\label{fig:cluster8}]{\fbox{\includegraphics[trim={20cm 10cm 20cm 10cm},clip, width=0.2\textwidth]{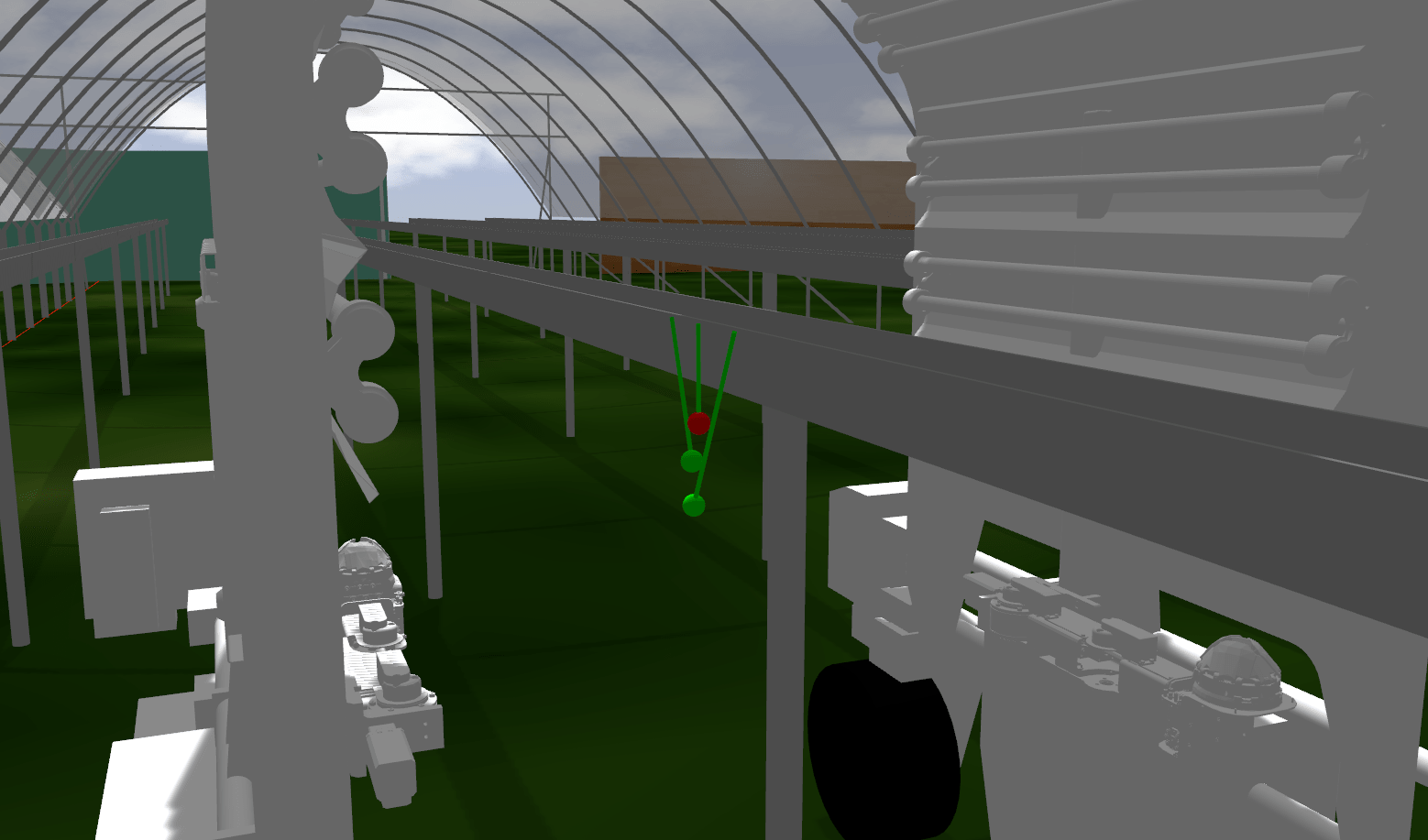}}}\hspace{50pt}
\subfloat[\label{fig:cluster9}]{\fbox{\includegraphics[trim={20cm 10cm 20cm 10cm},clip, width=0.2\textwidth]{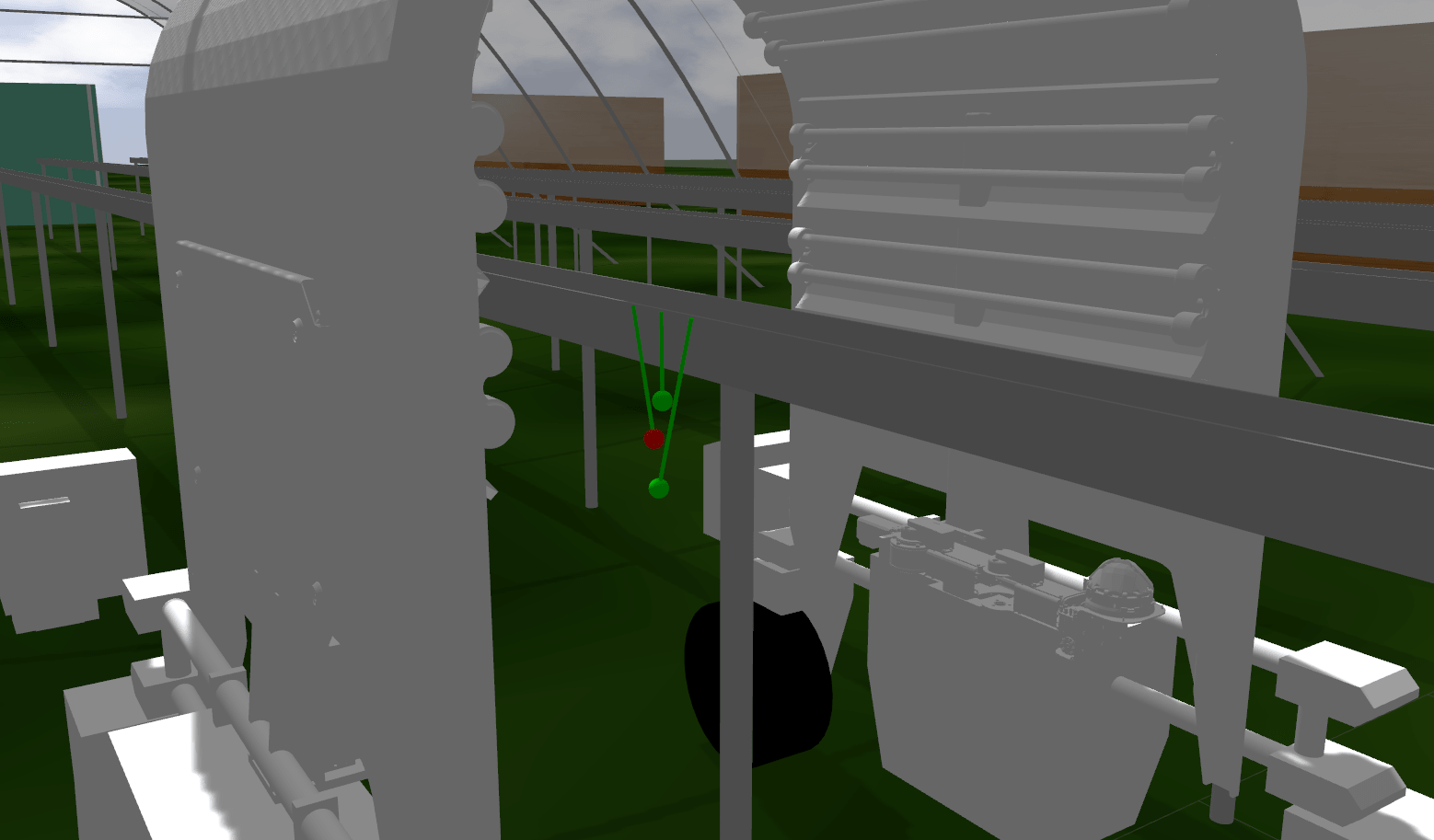}}}
\vspace{-0.2cm}\newline

\subfloat[\label{fig:traj7}]{\includegraphics[trim={4cm 1cm 2cm 2cm},clip, width=0.3\textwidth]{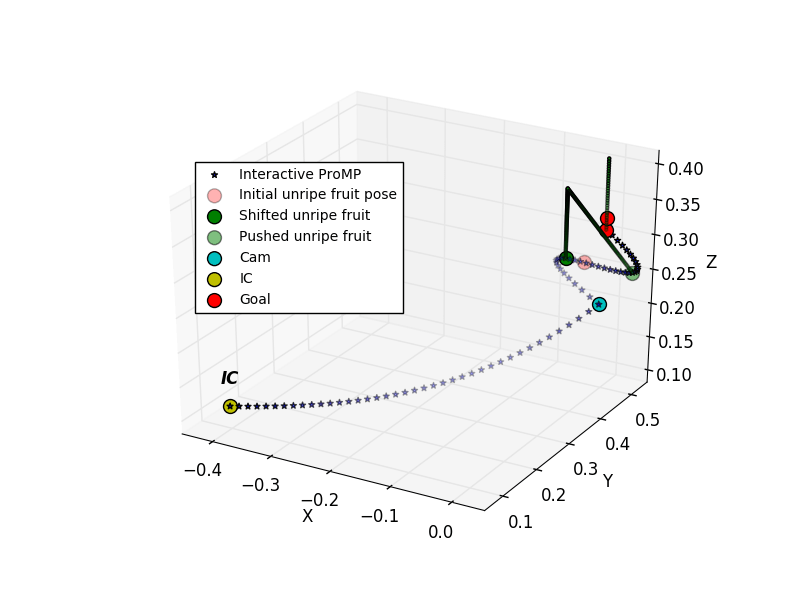}}\hspace{5pt}
\subfloat[\label{fig:traj8}]{\includegraphics[trim={4cm 1cm 2cm 2cm},clip, width=0.3\textwidth]{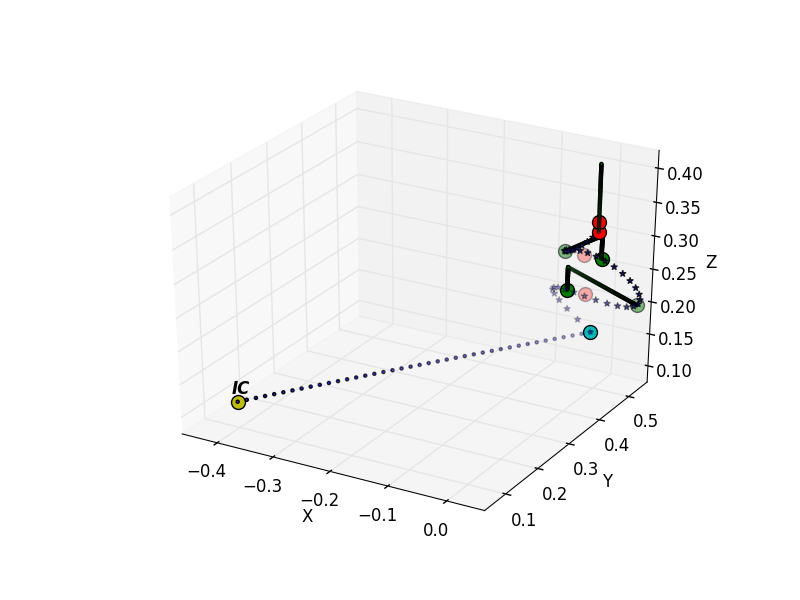}}\hspace{5pt}
\subfloat[\label{fig:traj9}]{\includegraphics[trim={4cm 1cm 2cm 2cm},clip, width=0.3\textwidth]{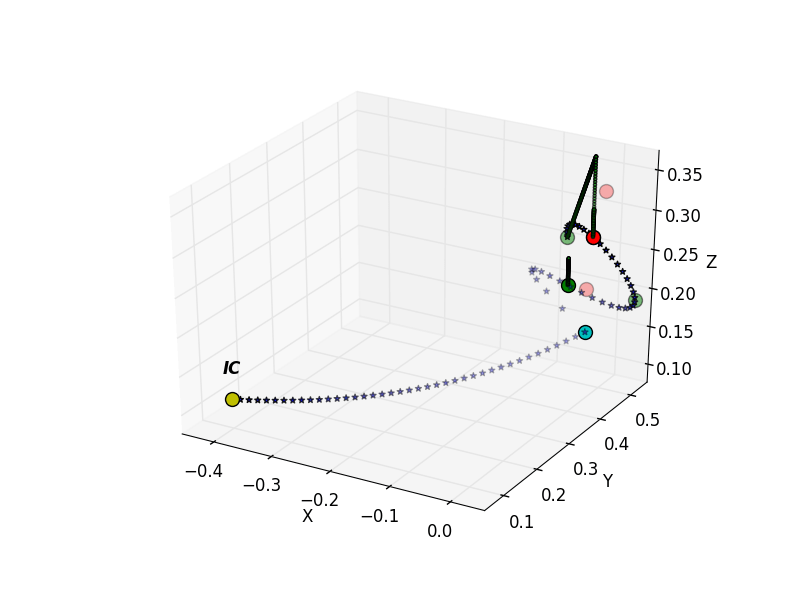}}
\vspace{-0.2cm}\newline

\subfloat[\label{fig:traj7sim}]{\includegraphics[trim={4cm 1cm 2cm 1.5cm},clip, width=0.3\textwidth]{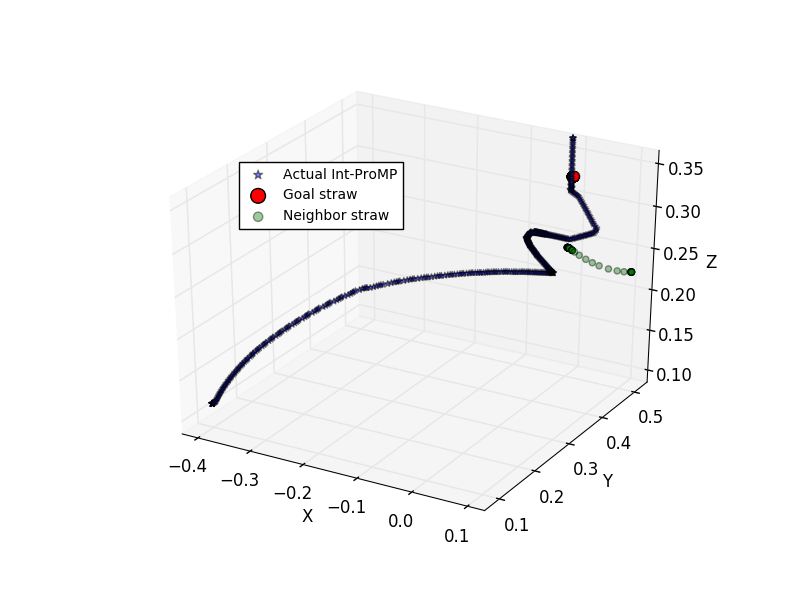}}\hspace{5pt}
\subfloat[\label{fig:traj8sim}]{\includegraphics[trim={4cm 1cm 2cm 1.5cm},clip, width=0.3\textwidth]{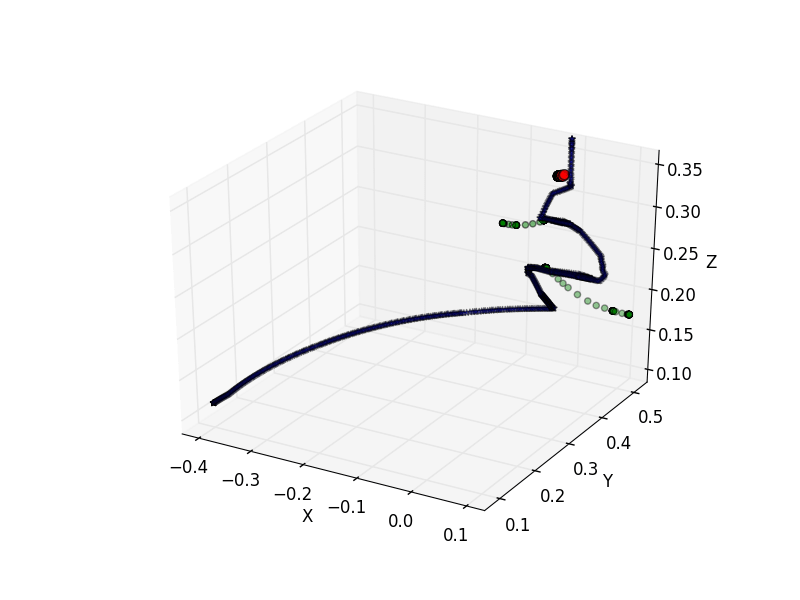}}\hspace{5pt}
\subfloat[\label{fig:traj9sim}]{\includegraphics[trim={4cm 1cm 2cm 1.5cm},clip, width=0.3\textwidth]{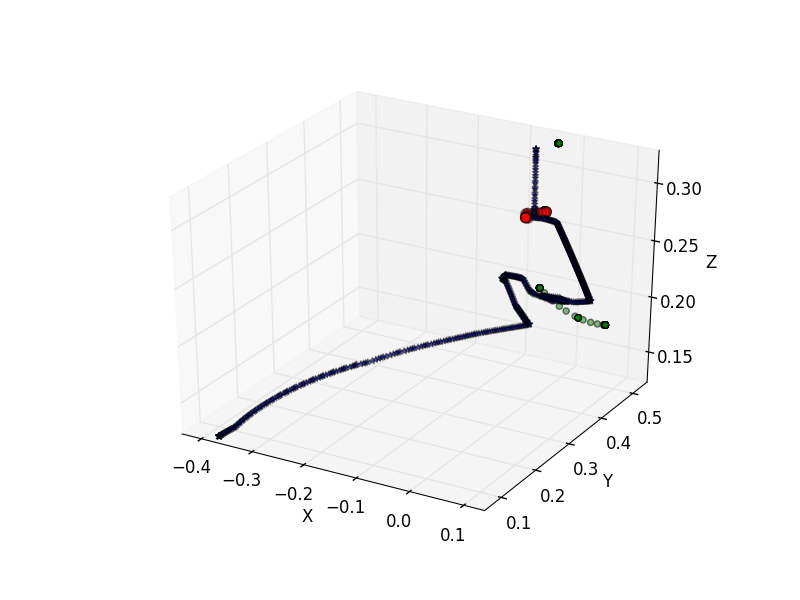}}
\caption{Probabilistic primitive-based pushing strategy to swallow a soft fruit from complex clusters: (a) cluster with target fruit (red) occluded by one stiff inclined stem and an unripe fruit (green), (b) cluster with target occluded by two stiff, unripe stem-fruit system, each from one side, (c) cluster with target fruit occluded by two stiff stem-fruit system, one from above and another one from below, (d-f) Interactive ProMP generated for the clusters (a-c): the target is conditioned at a point directly below it (representing the target radius, 1.5cm, plus a margin of 0.1cm). 
The pushed unripe fruit is connected to the shifted pose with an inclined green stem. The legend in (d) applies to (e-f) while legend in (g) applies to (h-i) too.
(g-i) show the actual end-effector trajectory, extracted from the simulation environment. Also, they illustrate the goal fruit trajectory and the neighbour unripe fruit trajectories.}\label{fig:trajComplexConfig}
\end{figure*}

We developed a simulation environment in \emph{Gazebo 8.0}, reported in fig. \ref{fig:sim1} and consisting of a polytunnel with parallel table tops. In addition to the configurations shown in fig. \ref{fig:clusters}, we built complex clusters (figs. \ref{fig:cluster8} - \ref{fig:cluster9}) out of the one shown in fig. \ref{fig:cluster7} and attached them to the table top. We sent Thorvald in the polytunnel and command the arm to follow a joint trajectory generated from the SPI planner. For the 3 simulated clusters, i.e figs. \ref{fig:cluster7} - \ref{fig:cluster9}, the SPI planner generates an interactive ProMP in the robot Cartesian space, shown in figs. \ref{fig:traj7} - \ref{fig:traj9} with a blue color, respectively. The joint space trajectories are first constrained to joint limits and then passed to the SCARA arm. They are monitored by an effort controller exposed to a follow joint trajectory action interface. We record the actual joint trajectories from the simulation and, with the use of the robot forward kinematics, we report in figs. \ref{fig:traj7sim} - \ref{fig:traj9sim} the actual end-effector trajectory and the trajectory of each strawberry in the scene. 

\textbf{Discussion:} In figs. \ref{fig:traj7} - \ref{fig:traj9}, every light pink sphere (without green stem) under or above the goal represents the initial pose of an unripe neighbour, accounted for with a horizontal shift in position. The latter shifted position is added to the conditioned set points and illustrated with a short stem. The shift accounts for an amount equal to $r_g^{max} + r_f^{max}$ where $r_f^{max}$ is an estimated maximum fruit radius; this is to align the end-effector with the pushing direction $\mathbf{u}_p$. We can see that the planner doesn't pass through the unripe fruit above the target in fig. \ref{fig:traj9} while the target sphere has an updated light green position generated by the planner to align its orientation vertically. For the scenario of fig. \ref{fig:traj8}, we computed the time for generating an I-ProMP. The mean computation time over 1000 samples is $0.19s$ with a standard deviation of $0.0022$.

In figs. \ref{fig:traj7sim} - \ref{fig:traj9sim}, it can be noticed that the actual end-effector trajectory passes through the unripe fruits, and reaches to the target (quasi-static position over time). We note that, the generated interactive primitive is not passed with multiple way-points to the robot joints before reaching $p_{bs}$, hence a larger difference between the generated one and the simulated one exists.
In table \ref{tab:metric}, a metric is retrieved to prove the occurrence of the pushing action. We consider a contact has occurred if the minimum distance traveled by the gripper frame with respect to a frame attached to fruit center, at the fruit altitude is: $0 < d_{f, ee}^{min} \leq (r_f^{max} + r_g^{max}) = 4.5cm$. For cases of $C_{I}$-neighbour$^1$ and neighbour$^1$ of configuration fig. \ref{fig:cluster9}, it results that the contact has occurred with the conical gripper surface below the frame attached to its vertex.  
This work is accompanied with an attached video reporting the simulation of each case scenario studied.

\begin{table}
\begin{center}
\begin{tabular}{ |p{2cm}|p{2cm}|p{2cm}|  }
 \hline
 & $h_{ee, f}^{min} (cm)$ &  $h_{ee, f}^{max} (cm)$\\
\hline
 \multicolumn{3}{|c|}{Configuration $C_{I}$} \\
\hline
$g_t$  & 2.12  &  2.12\\
Neighbour$^1$ &   4.95  & 4.95\\
Neighbour$^2$ &   1.22  & 1.22\\
\hline
 \multicolumn{3}{|c|}{Configuration fig. \ref{fig:cluster7}} \\
 \hline
$g_t$  & 1.368  &  1.369\\
Neighbour$^1$ &   2.784  & 5.107\\
\hline
 \multicolumn{3}{|c|}{Configuration fig. \ref{fig:cluster8}} \\
 \hline
$g_t$  & 0.7  &  0.75\\
Neighbour$^1$ &   0.6  & 9.06\\
Neighbour$^2$ &   1.6  & 5.98\\
\hline
 \multicolumn{3}{|c|}{Configuration fig. \ref{fig:cluster9}} \\
 \hline
$g_t$  & 2.46 &  2.49\\
Neighbour$^1$ &   6.1  & 8.3\\
 \hline
\end{tabular}
\end{center}
\caption{Metric for initial push action occurrence: $h_{ee,f}^{min}$ and $h_{ee,f}^{max}$ are the minimum and maximum distance, respectively, between gripper frame and $\{g_t \cup S_d^*\}$, when it reaches to the fruit altitude. }\label{tab:metric}
 \end{table}

\section{Conclusion}
In this work, we presented an interactive primitive-based planning strategy that features pushing actions in complex clusters. Although it can be generalized to different applications, the proposed approach targets a specific application, the one of robotic harvesting, and hence tackles the problem of picking occluded fruits. Occlusion results normally from the variety of grown clusters. In order to generate different degrees of non-linearity in the system behaviour, the planner learns two movement primitives from demonstrations, conditions the resulting primitive to pass through selective neighbours (movable obstacles), then finds the pushing direction of each of them and finally it augments them with an updated pose. We tested our approach on different complex cluster configurations in a simulated polytunnel using a SCARA robotic arm mounted on a mobile base. The strategy succeeded to reach the target in the different scenarios selected. As part of future work, the pushing planner will be tested on real field at the beginning of the coming strawberry season with an online closed loop feedback and primitive update. In addition, an optimization approach will be developed to take into consideration fixed obstacles and the minimum number of fruits to push.
 \label{sec:conc}

\section*{Acknowledgement}
The authors would like to thank Prof. Grzegorz Cielniak, Prof. Simon Pearsons, and SAGA Robotics Ltd together with its engineers Callum Small and Kristoffer Kskarsgard, for their inputs to this work and their support and contribution on the hardware mock-up platform (Thorvald Harvester) 

\bibliographystyle{IEEEtran}
\bibliography{IEEEabrv,references}

\end{document}